\documentclass[journal]{IEEEtai}

\usepackage[colorlinks,urlcolor=blue,linkcolor=blue,citecolor=blue]{hyperref}
\usepackage{subcaption} 
\usepackage{float}
\usepackage{multirow}
\usepackage[inkscapelatex=false]{svg}
\usepackage{lipsum}

\usepackage{color,array}
\usepackage{cite}

\usepackage{graphicx}
\usepackage{threeparttable}

\usepackage[table,xcdraw]{xcolor}

\begin{document}

\title{
Regularisation in neural networks: a survey and empirical analysis of approaches
}

\author{Christiaan P. Opperman, Anna S. Bosman, \IEEEmembership{Member, IEEE}, and Katherine M. Malan, \IEEEmembership{Senior Member, IEEE}
\thanks{C.P. Opperman and Anna S. Bosman are with the Department of Computer Science, University of Pretoria, South Africa (email: u17023239@tuks.co.za, anna.bosman@up.ac.za).}
\thanks{K.M. Malan is with the Department of Decision Sciences, University of South Africa, South Africa (email: malankm@unisa.ac.za)}
}

\markboth{This is a preprint version of a manuscript published in IEEE TAI: \url{https://doi.org/10.1109/TAI.2025.3644334}}
{Christiaan P. Opperman \MakeLowercase{\textit{et al.}}: Regularisation in neural networks: a survey and empirical analysis of approaches}

\maketitle

\begin{abstract}
Despite huge successes on a wide range of tasks, neural networks are known to sometimes struggle to generalise to unseen data. Many approaches have been proposed over the years to promote the generalisation ability of neural networks, collectively known as regularisation techniques. These are used as common practice under the assumption that any regularisation added to the pipeline would result in a performance improvement. In this study, we investigate whether this assumption holds in practice. First, we provide a broad review of regularisation techniques, including modern theories such as double descent. 
We propose a taxonomy of methods under four broad categories, namely: (1) data-based strategies, (2) architecture {strategies}, (3) training {strategies}, and (4) loss function strategies. {Notably}, we highlight the contradictions and correspondences between the approaches in these broad classes. Further, we perform an empirical comparison of the various regularisation techniques 
on classification tasks for ten numerical and image datasets applied to the multi-layer perceptron and convolutional neural network architectures. {Results show} that the efficacy of regularisation is dataset-dependent. For example, the use of a regularisation term only improved performance on numeric datasets, whereas batch normalisation improved performance on image datasets only. Generalisation is crucial to machine learning; thus, understanding the effects of applying regularisation techniques, and considering the connections between them is essential to the appropriate use of these methods in practice. 
\end{abstract}

\begin{IEEEImpStatement}
Generalisation is one of the most important properties of a neural network (NN), as it determines whether the NN would produce usable outputs outside of the lab setting. Current practices of improving NN generalisation via regularisation techniques are haphazard and unreliable. Our work sheds light on the interplay between the various existing regularisation techniques, and points out directions for future research that can streamline regularisation and make it more robust. We demonstrate that some popular methods of regularisation, such as dropout, {may be ineffective on smaller models and datasets}, and challenge the universality of regularisation techniques. We show that there is a strong need to tailor the regularisation approach to the task at hand, and that no one-size-fits-all solution exists.
\end{IEEEImpStatement}


\section{Introduction}

\IEEEPARstart{G}{eneralisation} ability is the performance of an artificial neural network (NN) on previously unseen data instances. Generalisation is one of the most important characteristics of NNs, as it determines the {applicability} of the final model {to real-world data}. Even though multiple theories have been proposed in an attempt to explain generalisation, {varying from linking generalisation to geometric properties of the loss landscape~\cite{hochreiter1997flat} to attributing generalisation to the size and structure of the neural architectures~\cite{belkin2019reconciling}}, this phenomenon is still not well understood~\cite{ronny2020Understanding,kawaguchi2022generalization,zhang2021understanding}. The generalisation theories have, however, led to the development of \textit{regularisation techniques}, which are commonly used to potentially improve generalisation. Some regularisation techniques were discovered incidentally, through research that investigated approaches to improve other properties of a NN, such as memory usage~\cite{blalock2020state}. Due to the lack of theoretical studies, regularisation techniques do not guarantee improved generalisation in NNs. Furthermore, the choice of which approach to use is a non-trivial design decision.

\begin{figure*}
    \centerline{\includegraphics[width=2\columnwidth]{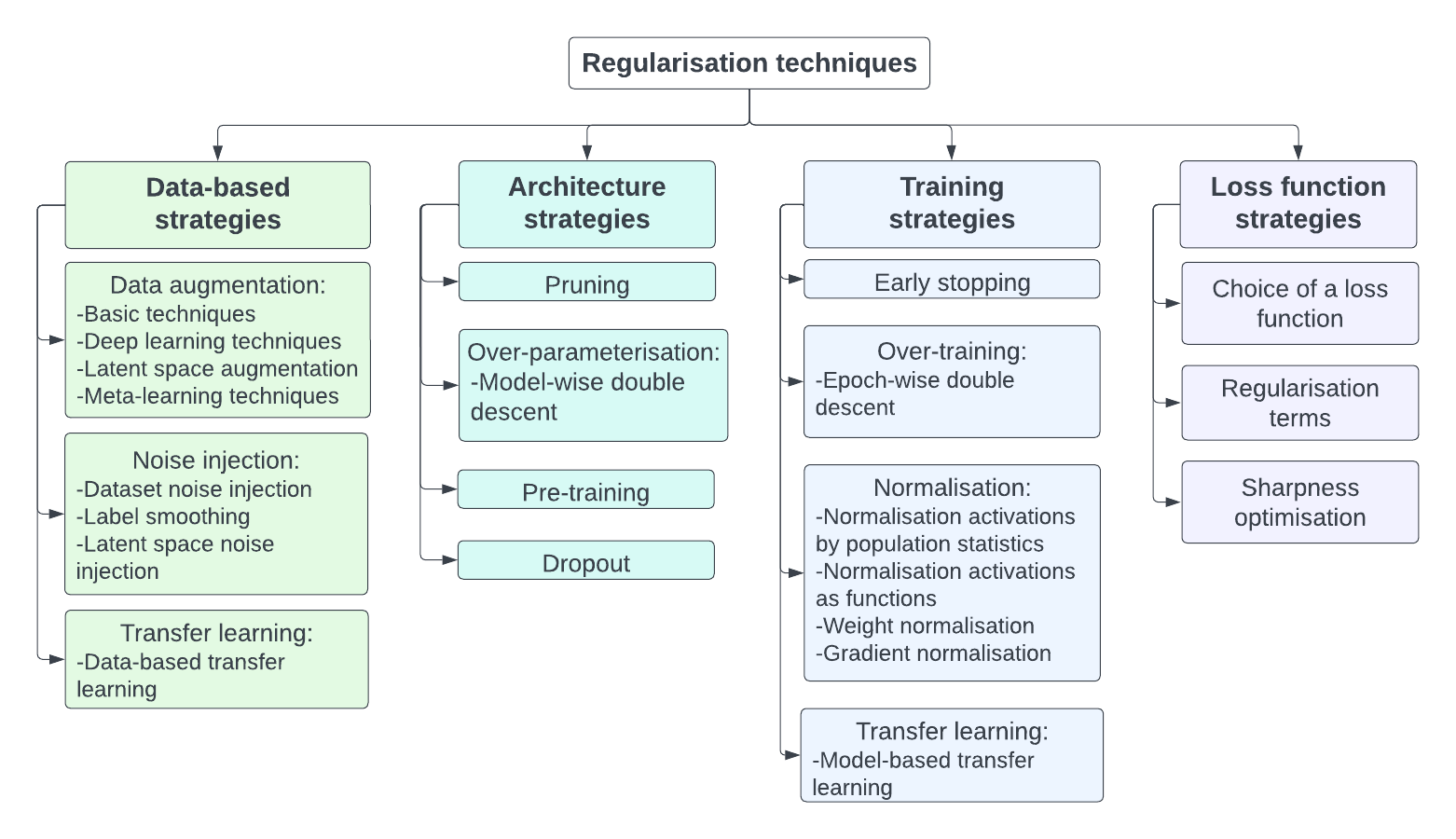}}
    \caption{A tree representation of the proposed taxonomy of regularisation methods.}
    \label{fig:mesh1}
\end{figure*}

In this paper, we survey the existing regularisation techniques, and make the following {specific contributions}:
\begin{enumerate}
\item We propose a taxonomy of regularisation techniques (summarised in Fig.~\ref{fig:mesh1}) under four broad categories, namely: data-based {strategies}, architecture {strategies}, training {strategies}, and loss function {strategies}. Each category is broken down into specific subcategories and approaches. This visual overview of the wide range of alternatives to promote generalisation provides a useful starting point for researchers and practitioners working in the field.
\item For each set of techniques proposed, we provide a brief description of the rationale underlying the general approach, and give references to sources that provide further details to guide researchers and practitioners on the implementation of the techniques.
\item We highlight \textit{contradictions} and \textit{correspondences} between regularisation techniques, suggesting that certain combinations of approaches may be in conflict with each other and should be used with caution until further studies provide guidance on ways to handle these apparent contradictions.
\item Finally, to evaluate the relative effectiveness of the regularisation techniques, we perform benchmark tests on two distinct NN architectures (multi-layer perceptron and convolutional neural network), on ten different datasets (numerical and image classification) for the following techniques: geometric transformation, SMOTE, weight perturbation, pruning, dropout, batch normalisation, layer normalisation, weight normalisation, and regularisation terms.
\item We observe that regularisation performance varies significantly across datasets and NN architectures, and in many cases, regularisation hinders rather than helps.  
\end{enumerate}

\textbf{Related work.} This paper is not the first survey of regularisation techniques. Kuka{\v{c}}ka et al.~\cite{kukavcka2017regularization} defined a high-level taxonomy of regularisation techniques based on which part of the NN the techniques affect: data, NN architecture, loss function, regularisation term, or optimisation. The taxonomy proposed in our paper is similar to Kuka{\v{c}}ka et al.'s~\cite{kukavcka2017regularization}, but {our survey includes additional strategies under loss functions} and, furthermore, discusses the contradictions and correspondences between these techniques. Further, Kuka{\v{c}}ka et al.'s~\cite{kukavcka2017regularization} review was carried out in 2017, and therefore excludes some of the most recent discoveries such as double descent and sharpness optimisation. Moradi et al.~\cite{moradi2020survey} performed a review of regularisation techniques in 2020, but did not provide a usable taxonomy, nor discuss the links and contradictions between the approaches. Similarly, in 2022, Tian and Zhang~\cite{tian2022comprehensive} performed a theoretical survey of regularisation techniques in machine learning, but did not provide a taxonomy or compare the approaches. Survey by Santos et al.~\cite{santos2022Avoiding} only considers more recent regularisation techniques, and focuses on image datasets and convolutional neural networks. Our paper puts prior work in a wider modern context of understanding generalisation, and provides a more holistic and practical view of regularisation.

{A traditional review approach was followed for the survey of techniques. We used the surveys discussed above as a basis to establish the main categories of the taxonomy. For each of the main categories, existing category-specific surveys were reviewed to identify subcategories. Further, we adopted a snowballing approach by mining the reference lists of the review articles, forward-tracking of citations in time, and performing target-specific searches on generalisation and regularisation techniques.} 

\textbf{Paper structure.} This paper is organised into two main sections. Firstly, Section~\ref{sec:taxonomy} describes the proposed taxonomy, providing a detailed view of the four categories of approaches that have been proposed to promote generalisation, and the contradictions and correspondences between them. 
Secondly, Section~\ref{sec:benchmarks} describes the benchmark tests that were performed on a subset of the regularisation techniques, and discusses the benchmarking results. Section~\ref{sec:conclusion} concludes the paper.

\section{Taxonomy}\label{sec:taxonomy}
Fig.~\ref{fig:mesh1} summarises the proposed taxonomy of regularisation techniques. Each of the main categories (data-based {strategies}, architecture {strategies}, training {strategies}, and loss function {strategies}) is discussed below in Sections~\ref{subsec:data_manip} to~\ref{subsec:loss_function_manip}. Section~\ref{subsec:discussion} provides an overview of the contradictions and correspondences between the families of regularisation techniques. 
\subsection{\textbf{Data-based {strategies}}}\label{subsec:data_manip}
Data-based regularisation {strategies} manipulate the training dataset or the latent representation of the data. The data is either increased in volume and/or variance, or used as a channel to transfer knowledge between NNs.
\subsubsection{\textbf{Data augmentation}}\label{subsubsection:data-augmentation}
A common way to improve the generalisation of a NN is to artificially expand the training dataset through data augmentation~\cite{shorten2019survey}. Expanding the dataset can introduce more diversity into the data and thus reduce the chance of overfitting. 

\textbf{Basic data augmentation techniques} 
apply functions (such as SMOTE~\cite{arslan2019smote}) or transformations (such as rotation~\cite{shorten2019survey}) directly to the data instances to create new variations of existing data instances. The new instances differ slightly from the original ones, but are typically given the same label. A promising recent alternative to this is mixup augmentation~\cite{zhang2018mixup}, which manipulates both the input features and the labels to create new data instances. A new data instance is created by randomly selecting two existing data instances, and linearly interpolating between their feature vectors and labels. Due to the fact that new data instances are constructed as linear combinations of the existing data instances, the augmented data adheres to the original data distribution. A survey of mixup techniques was performed by Jin et al.~\cite{jin2024survey}.

\textbf{Deep learning augmentation techniques} use generative deep learning models to create new data instances. The deep learning models that can be used for data augmentation include autoencoders~\cite{ohno2020auto}, generative adversarial networks (GAN)~\cite{shin2018medical}, and diffusion models~\cite{trabucco2023effective}. These models learn the data distribution of the input space, allowing the learnt distribution to be sampled for the purpose of generating new input data.

\par\textbf{Latent space augmentation techniques} augment data in the latent space, allowing for domain-agnostic augmentation~\cite{devries2017dataset}. Latent space refers to the internal, lower-dimensional representation of the original input data as learned by the NN. As such, both basic and deep learning augmentation techniques are applicable to the latent space representations.

\textbf{Meta-learning techniques} use machine learning techniques (meta-learners) to optimise machine learning models~\cite{huisman2021survey}. In the context of data augmentation, a meta-learner is used to determine the type and extent of data augmentation applied.
Influential examples of meta-learning techniques are auto augmentation~\cite{cubuk2018autoaugment}, where the meta-learning task is formulated using the reinforcement learning paradigm, and smart augmentation~\cite{lemley2017smart}, which uses a NN to learn the best way to combine existing samples to produce new ones.\par
The reader is referred to~\cite{shorten2019survey,xu2023Comprehensive,yang2023image} for surveys of augmentation methods for deep learning in the context of image data.

\subsubsection{\textbf{Noise injection}}\label{subsubsection:noise-injection}
It has been shown that noise injection (applying small perturbations to data) improves the generalisation of NNs~\cite{altarabichi2024rolling, an1996effects, holmstrom1992using}. 

The injected noise is randomly generated according to a specific distribution, such as a Gaussian distribution.

\textbf{Dataset noise injection}
involves the injection of noise into the features of a dataset~\cite{altarabichi2024rolling, holmstrom1992using}. Dataset noise injection depends heavily on the data being used to train the NN, as the type of data will determine how the noise is injected. For example, noise can be injected into an image dataset by adding a matrix of random values to an image~\cite{moreno2018forward}, while adding noise to a numerical dataset is as simple as adding random values to the features of the dataset. 

\par\textbf{Label smoothing} alters the labels of a dataset from hard labels to softer probability distribution labels~\cite{szegedy2015rethinking}. Various papers have shown that label smoothing not only improves NN regularisation but also improves the robustness of a NN against label noise~\cite{lukasik2020does,muller2019advances,pereyra2017regularizing}.

\par\textbf{Latent space noise injection} techniques add noise directly to the latent space. Places where noise has been injected, which resulted in improved generalisation, are the neuron activation signals~\cite{inayoshi2005improved,nori2023effectiveness}, the NN weights (i.e. weight perturbation)~\cite{goodfellow2016deep,khatami2020weight},  loss function~\cite{altarabichi2024rolling}, and the gradient weight updates (perturbed or masked gradients)~\cite{zhou2019toward, altarabichi2024rolling}. 

Altarabichi et al.~\cite{altarabichi2024rolling} provides a holistic overview of noise injection approaches.

\subsubsection{\textbf{Transfer learning}}\label{subsubsec:Transfer_learning}
Transfer learning is a set of techniques used to improve the training time and generalisation performance of a NN (the target) by transferring the knowledge gained from a previously trained NN (the source). 

If the source domain and target domain are not well-related to each other, the performance of a target NN will be negatively impacted, i.e. negative transfer will take place~\cite{rosenstein2005transfer}.
Zhuang et al.~\cite{zhuang2020comprehensive} 
classify transfer learning techniques into two categories, namely data-based (discussed below) and model-based (see Section~\ref{subsubsec:model-based}).
\par\textbf{Data-based transfer learning} adjusts and transforms the data to transfer the knowledge between NNs. Techniques include instance weighting strategies and feature transformation strategies~\cite{zhuang2020comprehensive}. In their survey on transfer learning, Zhuang et al.~\cite{zhuang2020comprehensive} summarise over 40 different transfer learning mechanisms and strategies and demonstrate the performance of transfer learning models through experimentation. More recently, Bao et al.~\cite{bao2024recent} surveyed 60 heterogeneous transfer learning techniques.
The instance weighting strategy~\cite{huang2006correcting} transfers the data instance directly from the source domain into the target domain. To mitigate negative transfer effects, each instance is assigned a weight indicating how related the instance is to the target domain instances, and only instances with weights equal to or above specified values are transferred. The feature transformation strategy~\cite{zhuang2020comprehensive} transfers knowledge by transforming the feature of a source domain into a new feature representation. This allows the transfer of the structure and other properties of the source domain data, reduces the difference between domain distributions, and finds corresponding features in the target domain data~\cite{zhuang2020comprehensive}. 

\subsection{\textbf{Architecture {strategies}}}\label{subsec:architecture_manip}
The architecture of a NN refers to the structural properties, such as the number of layers, the number of neurons in a layer, and the choice of activation function. This subsection discusses techniques that manipulate the architecture of a NN to improve generalisation. 

\subsubsection{\textbf{Pruning}}\label{subsubsection:pruning}
NN pruning is the act of reducing the size of a NN by removing unnecessary parameters. It has been observed that parameter reduction often leads to improved generalisation, even though the primary goal of pruning is to reduce the computational cost of NN training and deployment~\cite{bartoldson2020generalization,jin2022pruning}. A  survey by Blalock et al.~\cite{blalock2020state} provides an overview of pruning approaches, as well as an open-source framework for the evaluation and empirical analysis of different pruning methods. Blalock et al.~\cite{blalock2020state} group pruning techniques based on sparsity structure, scoring, scheduling, and fine-tuning. 
Similarly to Blalock et al.~\cite{blalock2020state}, Zhu et al.~\cite{zhu2025comprehensive} in their survey of pruning grouped the techniques based on pruning granularity (i.e., sparsity structure) and pruning time (i.e., scheduling). 
 
The sparsity structure refers to the parameters affected by a pruning technique. Pruning individual parameters, or weights, is referred to as unstructured pruning~\cite{yang2021comparative}, while pruning related groups of parameters is referred to as structured pruning~\cite{yang2021comparative, he2023structured}. Parameter groups used in structured pruning can consist of entire neurons, filters, channels, or layers.\par
Scoring refers to the method of quantifying the importance of parameters~\cite{blalock2020state}. The parameters are most commonly scored based on their individual absolute value, trained importance coefficients, or contribution to NN activations and/or gradients. Parameter scores can be compared either to a baseline value, the scores of other parameters in a local substructure (e.g. within a layer of a NN)~\cite{han2015learning}, or the scores of other parameters in the entire NN~\cite{lee2018snip}.\par
Further choices that greatly impact pruning are when such pruning should occur (referred to as scheduling), and the number of parameters that should be pruned (referred to as scaling). Pruning can occur before the training of a NN starts~\cite{lee2018snip}, at multiple times during the training of a NN (referred to as iterative pruning)~\cite{tan20a}, or at the end of NN training~\cite{han2015learning}. If pruning only occurs at the end of training, fine-tuning is required. The number of parameters to prune can be determined by a specific value~\cite{li2016pruning}, a fixed percentage~\cite{mondal2022adaptive}, or a score threshold~\cite{han2015learning}. 

\subsubsection{\textbf{Over-parameterisation}}\label{subsubsection:over_parameterisation}
In statistical learning theory, a basic tenet is that the more complex a model is, the more likely it is to overfit~\cite{hastie2009elements}. Deep NNs tend to follow this tenet up to a point, after which the model becomes less likely to overfit -- i.e., as the complexity of the model, $N$, approaches the complexity of the dataset, $n$, the generalisation accuracy of the model decreases, but once $N$ surpasses the \textit{interpolation threshold}, i.e., the complexity necessary for the NN to perfectly fit the training dataset, the generalisation accuracy of the model begins to improve again as $N$ increases~\cite{belkin2019reconciling}. This tendency is known as the {\it double descent} and has already been studied as far back as 2000~\cite{caruana2000overfitting}. Nakkiran et al.~\cite{nakkiran2021deep} distinguish between two types of double descent, namely model-wise double descent (discussed below) and epoch-wise double descent (see Section~\ref{subsubsubsec:epoch_wish}). Both describe how a model can surpass the interpolation threshold.
\par\textbf{Model-wise double descent} occurs when more neurons are added to each layer of the NN, or when extra hidden layers are added, until the interpolation threshold is surpassed. NNs which follow model-wise double descent are referred to as over-parameterised NNs. Belkin et al.~\cite{belkin2020two} studied the number of parameters that are needed for model-wise double descent to occur, and found that, in general, the number of parameters $N_p$ should be close to or exceed the number of samples $n_p$ that are used for training.

\subsubsection{\textbf{Pre-training}}\label{subsubsection:pretraining}
Pre-training techniques involve training simple models as a preliminary task on the way to training more complex models~\cite{goodfellow2016deep}. However, it has been shown that pre-training techniques have other benefits, such as better generalisation and lower variance in the final testing error~\cite{erhan2009difficulty}. Pre-training techniques can be grouped into three categories: supervised, unsupervised, and self-supervised.\par
Supervised pre-training techniques progressively add hidden layers directly to a model trained on a supervised learning task~\cite{han2018batch}. Unsupervised pre-training techniques build a deep unsupervised autoencoder model, trained to reconstruct the inputs, after which the decoder part is discarded, and a supervised output layer is added to the encoder part in order to enable the NN to solve the target problem~\cite{erhan2009difficulty}. Self-supervised pre-training relies on the contrastive learning paradigm, where the model receives augmented versions of the input data and is trained to recognise the similarity between the original and the augmented inputs~\cite{reed2022self}. Chen et al.~\cite{chen2020simple} show that self-supervised pre-training techniques outperform more traditional supervised and unsupervised pre-training techniques for image classification and provide a software framework for implementation. 

It is worth noting that self-supervised pre-training gave rise to the so-called \textit{foundation models}, i.e. NNs typically trained on vast amount of data in a self-supervised manner, often in different modalities such as both image and text, for the purpose of discovering an internal representation of the data that can be easily adapted to various downstream tasks such as classification or regression~\cite{awais2025, zhou2024comprehensive}.

\subsubsection{\textbf{Dropout}}\label{subsubsection:dropout}
Dropout was introduced by Hinton et al.~\cite{hinton2012improving} in 2012. During dropout, a random selection of input and hidden neurons are temporarily deactivated for each batch during training. The neurons are reactivated at the end of each batch, and a different randomly chosen set of neurons is deactivated. This process causes each neuron not to rely on other neurons, resulting in more robust models with improved generalisation. Multiple variations of dropout have been proposed; see Li et al.'s~\cite{yangkun2023survey} survey.

\subsection{\textbf{Training {strategies}}}\label{subsec:training_manip}
The training process of a NN has a large impact on its generalisation. This subsection summarises techniques that have been used to manipulate the training process of a NN with the purpose of promoting generalisation.

\subsubsection{\textbf{Early stopping}}\label{subsubsection:early-stopping}
Early stopping of the NN training is based on the observation that the testing error together with the training error reduces up to a point, after which the training and testing errors diverge, and generalisation deteriorates despite improvements in the training error~\cite{morgan1989generalization}. Early stopping refers to the set of stopping criteria used to automatically determine when the training of the NN should be halted. If the halting criterion is based on the testing error, it is called early stopping with cross-validation, while if the halting criterion is based on the training error, it is called early stopping without cross-validation \cite{ferro2023early}.
\par
Early stopping with cross-validation was first introduced by Morgan and Bourlard~\cite{morgan1989generalization} in 1989, who recommended using an independent test set
during the course of training to monitor generalisation performance. As such, the most commonly used early stopping methods are to stop training when a set test error threshold is reached, or when the (average) test error does not improve after a set number of epochs~\cite{prechelt1998automatic}. More complex criteria are also used in the literature, such as stopping when the training error becomes noisy or using a combination of different criteria~\cite{lodwich2009evaluation}.

Training can also be stopped through the use of a signal-to-noise-ratio figure (SNRF) in the context of function approximation, where SNRF is calculated on the training set and estimates the amount of information in the error signal still not learned by the NN~\cite{liu2008optimized}.
Statistical methods have also been used to determine when to stop, such as the approach followed by Iyer et al.~\cite{iyer2000novel}, where the performance-to-cost ratio of the network is calculated and used to determine when to stop training. 

\subsubsection{\textbf{Over-training}}\label{subsubsection:over_training}
Section \ref{subsubsection:over_parameterisation} discussed the concept of the interpolation threshold of a NN and double descent. Like model-wise double descent, epoch-wise double descent~\cite{nakkiran2021deep} is a phenomenon where the interpolation threshold is surpassed. 
\par\textbf{Epoch-wise double descent}\label{subsubsubsec:epoch_wish} occurs when sufficiently large models with long enough training periods experience a double descent of the testing error~\cite{nakkiran2021deep}. The phenomenon manifests in an initial decrease in testing error, followed by an increase in testing error near the interpolation threshold (conventionally seen as overfitting), followed by a subsequent period of decreasing test error. Therefore, epoch-wise double descent enables a NN to correct overfitting if the NN is allowed to train for a sufficient period. Nakkiran et al.~\cite{nakkiran2021deep} experiment with different learning tasks, architectures and optimisation methods and hypothesise when epoch-wise double descent will occur.

\subsubsection{\textbf{Normalisation}}\label{subsubsection:normalisation}
Normalisation is widely used to improve various metrics of a NN, such as training speed and generalisation. Huang et al.~\cite{huang2023normalization} provide an extensive review of normalisation techniques for deep NNs and highlight four main approaches used for improving NN performance: normalising activations by population statistics, normalising activations as functions, normalising weights, and normalising gradients.\par
\textbf{Normalising activations by population statistics}
considers the neuron activations across the NN as a single population. This population is then used to calculate statistics such as the mean and standard deviation of the activations over the dataset, and normalise (e.g. center, scale, etc.) the activations accordingly. The statistics are considered constant during the training process, which is an important disadvantage of these types of approaches, since they cause instabilities during training due to inaccurate estimations of population statistics~\cite{wiesler2014mean, huang2023normalization}. 
\par
\textbf{Normalising activations as functions} uses the population statistics of the activations over mini-batches. Batch normalisation was the first of these techniques~\cite{ioffe2015batch}. Normalising activations as functions no longer assumes population statistics to be constant, which addresses the drawbacks of normalising by population statistics~\cite{huang2023normalization}. 
Normalising activations as functions uses either a single type of normalisation, or a combination of different types of normalisation (e.g. channel-wise and layer-wise normalisation)~\cite{huang2023normalization}.\par
\textbf{Weight normalisation} during training is inspired by the practice of normalising weights during initialisation, and is commonly achieved by enforcing constraints on the weights of each layer~\cite{huang2023normalization}. Normalising the weights during training implicitly normalises the activations, which has been observed to improve the generalisation of a NN~\cite{huang2018orthogonal,huang2017centered,ozay2018training}. Weight normalisation tends to provide subpar generalisation gains when compared to normalising activations~\cite{gitman2017comparison}, and combinations of weight and activation normalisation tend to be more effective in practice~\cite{huang2023normalization}.
\par\textbf{Gradient normalisation} techniques attempt to alleviate the effect of ill-conditioning commonly observed in NNs. As such, gradient normalisation does not affect the search space, but rather alters the trajectory of a gradient-based optimiser~\cite{huang2023normalization}. Yong et al.~\cite{yong2020gradient} propose gradient centralisation, which centralises the gradients to have a zero mean, and effectively combines it with activation normalisation.

\subsubsection{\textbf{Transfer learning}}\label{subsubsec:model-based}
As stated in Section~\ref{subsubsec:Transfer_learning}, transfer learning techniques aim to transfer knowledge between a source learning task and a target learning task. Other than the data-based perspective, transfer learning can also be approached from the model perspective.
\par\textbf{Model-based transfer learning}~\cite{zhuang2020comprehensive} techniques adjust and transform the model parameters to transfer the knowledge between NNs. Example strategies include the following. Parameter control strategies share the parameters (such as weights) of a NN trained on the source domain with a NN trained on the target domain~\cite{zhuang2020comprehensive}. Model ensemble strategies train multiple NNs on different source domains and combine them into one model, which can solve the target learning task~\cite{ghorbanali2022ensemble}. Deep learning techniques can be used to construct transfer learning models and techniques~\cite{zhuang2020comprehensive}. Traditional deep learning techniques use methods such as autoencoders with shared weights to learn a bridging representation between two domains~\cite{zhuang2015supervised}. Adversarial deep learning techniques are based on the assumption that effective domain transfer is achieved when the extracted features do not discriminate between the source and target domains~\cite{ganin2016domain}. As such, a deep model can be trained in an adversarial fashion to learn a transferable feature representation~\cite{ganin2016domain}.
See Zhuang et al's~\cite{zhuang2020comprehensive} survey for further details and examples of applications of transfer learning.

\subsection{\textbf{Loss function {strategies}}}\label{subsec:loss_function_manip}
The loss function of a NN measures the extent to which the NN fails to fit a given dataset, and serves as a guide to a NN training algorithm. A number of studies have shown that generalisation can be improved by manipulating the loss function. This subsection discusses techniques that manipulate the loss function through the choice of a loss function, the use of a regularisation term, and sharpness optimisation.

\subsubsection{\textbf{Choice of a loss function}}
Two popular choices of loss functions are the quadratic error function (mean squared error) and the cross-entropy error (log likelihood). The choice of the loss function can impact the generalisation of a NN, since different loss functions are better suited to specific domains~\cite{gao2022loss}. Golik et al.~\cite{golik2013cross} theoretically showed that entropic error should result in better generalisation than quadratic error. However, Bosman et al.~\cite{bosman2020visualising} empirically showed that the quadratic error loss function was more resilient to overfitting than cross-entropy. The hybridisation of quadratic error and entropic error has also been shown to have improved generalisation potential over the individual loss functions~\cite{dickson2021hybridised}. Gonzalez and Miikkulainen~\cite{gonzalez2020improved} used evolutionary algorithms to evolve loss functions that provided better generalisation than cross-entropy for image classification tasks.

\subsubsection{\textbf{Regularisation terms}}\label{subsubsection:regularisation}
One of the most commonly used approaches to improve generalisation is to add a regularisation or penalty term to the loss function. Unlike the loss function, a regularisation term is typically independent of the target, which allows it to penalise the NN based on other properties, such as complexity. Minimising NN complexity alongside with the error is argued to yield models that only retain the essential parameters, and as such are less likely to overfit. Most commonly used penalty functions are $L_1$ and $L_2$ regularisation~\cite{moradi2020survey,goodfellow2016deep}, which in the context of NNs correspond to the sum of absolute values of the weights ($L_1$), and the sum of squares of the weight values ($L_2$). The $L_2$ regularisation is commonly known as \textit{weight decay}, however, other quantifiable properties of NNs besides weights may be used to construct a regularisation term~\cite{kukavcka2017regularization}, such as the gradient of the activation function with respect to the weights~\cite{hochreiter1994simplifying} and the gradient of the activation function with respect to the inputs~\cite{rifai2011higher}.\par

\subsubsection{\textbf{Sharpness optimisation}}\label{subsubsection:sharpness_aware_minimization}
The loss function of a NN generates a high-dimensional, complex and largely non-convex loss landscape with numerous minima. Multiple theoretical and empirical studies~\cite{bisla2022lowpass, bosman2020loss, bosman2020visualising, chaudhari2019entropy, dziugaite2017computing, petzka2021Relative} have investigated the connection between the characteristics of a loss landscape and the generalisation ability of the NN, and found that flatter minima often lead to better generalisation. Sharpness optimisers attempt to take advantage of these findings by minimising not only the loss value of a NN, but also the sharpness of the minima. There are two main approaches in this category, namely stochastic weight averaging (SWA)~\cite{izmailov2019averaging} and sharpness-aware minimisation (SAM)~\cite{foret2021sharpness}. SWA techniques are based on the fact that during training, a NN will traverse flat minima, but rarely reach the central point of the minima. SWA techniques correct this by averaging the weights which are in the minima together. SAM techniques find flat minima by minimising the maximum loss value in the neighbourhood of the current training step. Kaddour et al.~\cite{kaddour2022when} performed a comparison of SWA and SAM techniques on various datasets and problem domains, and found that the generalisation performance of SWA and SAM depends on the architecture, dataset and problem domain of the NN.

\subsection{\textbf{Discussion of contradictions and correspondences}}\label{subsec:discussion}
While compiling the taxonomy of regularisation techniques, numerous contradictions were found between the families of techniques. 
\begin{itemize}
\item \textbf{Early stopping and over-training}: 
Over-training~\cite{nakkiran2021deep} (Section~\ref{subsubsection:over_training}) indicates that training an overparameterised NN for longer allows the NN to correct overfitting over time, which can result in good generalisation performance. In contrast, early stopping (Section~\ref{subsubsection:early-stopping}) prescribes halting the training when the first signs of overfitting are observed. As such, early stopping is in direct conflict with over-training, which is acknowledged by Nakkiran et al.~\cite{nakkiran2021deep}. Should overfitting not be allowed, or should we let the NN attempt self-correction? Further studies are required to understand when epoch-wise double descent occurs, and whether early stopping may still be applicable after self-correction happens. Would a NN overfit again after correcting itself if left training for long enough?
\par
\item \textbf{Double descent and data augmentation or noise injection}: Data augmentation (Section~\ref{subsubsection:data-augmentation}) and noise injection (Section~\ref{subsubsection:noise-injection}) techniques are used to create new training data instances, i.e., increase the dataset size with the intention of improving generalisation performance. Nakkiran et al.~\cite{nakkiran2021deep} in their study of double descent linked to over-parameterisation (Section~\ref{subsubsection:over_parameterisation}) and over-training (Section~\ref{subsubsection:over_training}) discovered that the effect of a larger dataset on generalisation is correlated with the NN architecture size, and in certain cases, when a NN is not sufficiently under- or over-parameterised, more data may actually hurt rather than improve generalisation. Future research needs to investigate this relationship further, and provide tangible guidelines on how much to augment the data based on the chosen architecture dimensionality.
\item \textbf{Pruning and over-parameterisation}: Pruning (Section~\ref{subsubsection:pruning}) removes parameters from a trained NN to create a sparser network. Sparse NNs are not only more computationally efficient, but can often outperform dense NNs in terms of generalisation~\cite{bartoldson2020generalization,jin2022pruning}. On the other hand, over-parameterisation (Section~\ref{subsubsection:over_parameterisation}), i.e., using a NN model with the number of parameters that significantly exceeds the number of data points, was shown to simplify the learning task and induce the double descent behaviour, also yielding generalising solutions. Pruning seems to be in direct conflict with over-parameterisation. While some hypotheses were put forward to explain this contradiction~\cite{frankle2018lottery},  further research is required to understand why both these families of techniques can improve generalisation, and how they can be combined in the most effective way. Is pruning of an over-parameterised model more effective than the pruning of a NN that is not over-parameterised? \par
\end{itemize}
Besides the contradictions listed above, the following correspondences and similarities between the families of techniques were also noted that need further investigation:
\begin{itemize}
\item \textbf{Dataset noise injection and data augmentation}: These families of techniques, discussed in Subsections~\ref{subsubsection:data-augmentation} and~\ref{subsubsection:noise-injection}, both create new data instances, but differ in how the instances are created. Shorten and Khoshgoftaar~\cite{shorten2019survey} classified dataset noise injection as a subcategory of data augmentation. Indeed, overlaying data with noise can be seen as sampling new data instances from a distribution centred around a given data point. As such, using a unified terminology may be conducive to research in this field.
\item \textbf{Dataset noise injection and regularisation term}: Dataset noise injection (Section~\ref{subsubsection:noise-injection}) indicates that training a NN with noisy data improves generalisation performance. Bishop~\cite{bishop1995training} argued that certain regularisation terms (Section~\ref{subsubsection:regularisation}) added to the loss function can be shown to be equivalent to training with noisy data. As such, Bishop~\cite{bishop1995training} suggested using a regularisation penalty term as an alternative to noise injection. This relationship can be explored further, in the context of modern NN architectures and various penalty functions. 
\item \textbf{Dropout and pruning}: These families of techniques both alter the architecture of a NN by removing NN parameters, where dropout (Section~\ref{subsubsection:dropout}) removes parameters temporarily, and pruning (Section~\ref{subsubsection:pruning}) removes parameters permanently. A clear synergy between the approaches suggests that dropout methods may be usable in a pruning setting. Some studies combining dropout and pruning already exist~\cite{gomez2019learning}, but the topic remains largely underexplored.
\item \textbf{Transfer learning and pre-training}: The aim of transfer learning techniques (Subsections~\ref{subsubsec:Transfer_learning},~\ref{subsubsec:model-based}) is to transfer knowledge between domains and NN models. Greedy pre-training techniques (Section~\ref{subsubsection:pretraining}) construct a NN layer by layer to extract more meaningful latent representations. The constructive nature of pre-training suggests that newly added layers may be trainable on a new domain. In fact, training a fully connected layer on a target domain with a convolutional NN backbone trained on another domain is common practice in transfer learning~\cite{pan2009survey}, which can be seen as a variant of layer-wise pre-training. Further investigations of layer-wise pre-training in the context of transfer learning may prove fruitful.
\item \textbf{Pre-training and pruning}: Chen et al.~\cite{chen2021lottery} found that pre-trained NNs contain subnetworks similar to those found by iterative pruning in both structure and performance. Based on these findings, can it be concluded that such subnetworks will also be found in NNs created by other families of regularisation techniques?
\end{itemize}
The various correspondences and contradictions between the families of regularisation techniques show that regularisation techniques have been well-researched individually. However, further research is needed with regard to the interactions between these techniques and the underlying mechanics governing generalisation.

\section{Empirical analysis}\label{sec:benchmarks}
To complement the theoretical review provided in this study, we performed an empirical analysis of a representative selection of regularisation techniques. The efficacy of various regularisation methods was compared on classification tasks for both tabular and image data, using both multilayer perceptron and convolutional neural network architectures. This section provides the details and results of the empirical analysis. 

\subsection{\textbf{Datasets}}\label{subsec:Datasets}
The empirical analysis was performed in the problem domains of image and numerical data classification. Five datasets with different sizes, features, number of instances, {whether balanced or not}, and number of classes, were selected for each problem domain. {Only image datasets that are immune to potentially label-breaking data augmentation techniques, such as rotation, were selected.} See Table~\ref{tab:Datasets} for a summary of the datasets used. 

\begin{table*}[]
    \centering
    \caption{Summary of the datasets used in the empirical analysis}
    \label{tab:Datasets}
    \begin{tabular}{|l|l|l|r|r|l|l|}
        \hline
        \multicolumn{1}{|c|}{\textbf{Name}}                                               	& \multicolumn{1}{c|}{\textbf{Type}} & \multicolumn{1}{c|}{\textbf{Description of Features}}														& \multicolumn{1}{c|}{\textbf{\begin{tabular}[c]{@{}c@{}}No. of \\ Classes\end{tabular}}} & \multicolumn{1}{c|}{\textbf{\begin{tabular}[c]{@{}c@{}}No. of \\ Instances\end{tabular}}}	& \multicolumn{1}{c|}{\textbf{Balanced}}	& \multicolumn{1}{c|}{\textbf{Reference}}	\\ \hline
        Diabetes                                                                          	& Numeric                            & \begin{tabular}[c]{@{}l@{}}14 binary features and 7 numeric features\end{tabular}				& 2                                                                                          & 70 692																						& Yes                                       & ~\cite{Kahn2023Diabetes}				\\ \hline
        Liver Cirrhosis                                                                   	& Numeric                            & \begin{tabular}[c]{@{}l@{}}5 binary features and 12 numeric features\end{tabular}				& 3                                                                                          & 25 000																						& No                                        & ~\cite{Dickson2023Cirrhosis}			\\ \hline
        \begin{tabular}[c]{@{}l@{}}MAGIC Gamma Telescope\end{tabular}                  	& Numeric                            & 10 real features																				& 2                                                                                          & 19 020																						& No                                        & ~\cite{Bock2007Magic}					\\ \hline
        Mfeat pixel                                                                       	& Numeric                            & 19 numeric features																				& 10                                                                                         & 2 000																						& Yes                                       & ~\cite{Olson2017PMLB}					\\ \hline
        White Wine Quality                                                                	& Numeric                            & \begin{tabular}[c]{@{}l@{}}2 numeric features and 9 real features\end{tabular}				& 7                                                                                          & 4 898																						& No                                        & ~\cite{Cortez2009ModelingWP}			\\ \hline
        \begin{tabular}[c]{@{}l@{}}30 Types of Balls Updated\end{tabular}              	& Images                             & \begin{tabular}[c]{@{}l@{}}224 x 224 images in 3 colour channels\end{tabular}	& 30                                                                                         & 3 745																						& Yes                                       & ~\cite{Piosenka2023Types}				\\ \hline
        Bean Leaf Lesions                                                                 	& Images                             & \begin{tabular}[c]{@{}l@{}}500 x 500 images in 3 colour channels\end{tabular}	& 3                                                                                          & 1 167																						& No                                        & ~\cite{Marquis2020Makerere}			\\ \hline
        Cifar10                                                                           	& Images                             & \begin{tabular}[c]{@{}l@{}}32 x 32 images in 3 colour channels\end{tabular}	& 10                                                                                         & 60 000																						& Yes                                       & ~\cite{Krizhevsky2009Learning} 		\\ \hline        
        Fashion MNIST                                                                 	  	& Images                             & \begin{tabular}[c]{@{}l@{}}28 x 28 images in 1 colour channels\end{tabular}	& 10                                                                                         & 70 000																						& Yes										& ~\cite{Han2017Fashion}				\\ \hline	
        \begin{tabular}[c]{@{}l@{}}Nike, Adidas and Converse Shoes\end{tabular}	& Images                             & \begin{tabular}[c]{@{}l@{}}240x240 images in 3 colour channels\end{tabular}	& 3                                                                                          & 824																							& Yes                                       & ~\cite{iron2022nike}				\\ \hline
    \end{tabular}%
\end{table*}

\subsection{\textbf{Regularisation techniques}}\label{subsec:RegularisationTechniques}
The performance of a subset of the regularisation techniques described in Section~\ref{sec:taxonomy} was analysed on a range of different datasets. 
To illustrate the differences in the categories of regularisation techniques, the subset contains at least one of the techniques listed in each category. The popularity and complexity of each technique were also taken into account when making the selection of the subset of techniques, namely:
\begin{itemize}
    \item Geometric transformation (Section~\ref{subsubsection:data-augmentation}), implemented using PyTorch TorchVision's {\tt RandomRotation} with degrees set to be between $-180^\circ$ and $180^\circ$. 
    \item SMOTE (Section~\ref{subsubsection:data-augmentation}), implemented using {\tt imbalanced-learn}'s {\tt SMOTE} for datasets consisting only of numeric features, and {\tt SMOTENC} for datasets consisting of numeric and categorical features. 
    \item Weight perturbation (Section~\ref{subsubsection:noise-injection}), implemented by adding small random values to each weight of each linear and convolutional layer.
    \item Pruning (Section~\ref{subsubsection:pruning}), implemented using PyTorch's {\tt l1\_unstructured} pruning function called on specific epochs.     
    \item Dropout (Section~\ref{subsubsection:dropout}), implemented using PyTorch's {\tt Dropout1d} for fully connected layers, and {\tt Dropout2d} for convolutional layers.
    \item Batch normalisation (Section~\ref{subsubsection:normalisation}), implemented using PyTorch's {\tt BatchNorm1d} for fully connected layers, and {\tt BatchNorm2d} for convolutional layers.
    \item Layer normalisation (Section~\ref{subsubsection:normalisation}), implemented using PyTorch's {\tt LayerNorm} function. 
    \item Weight normalisation (Section~\ref{subsubsection:normalisation}), implemented using PyTorch's {\tt weight\_norm} function to wrap all linear and convolutional layers of the NN.
    \item Regularisation terms (Section~\ref{subsubsection:regularisation}), implemented by applying an $L_2$ regularisation term to the loss function. 
\end{itemize}
 The SMOTE technique was only applied to the numeric datasets, and could not be applied to the white wine quality dataset due to insufficient instances in one of its classes. The geometric transformation technique was only applied to the image datasets. Python code implementation used for the experiments is available at (URL to be provided after review).

\subsection{\textbf{Environment}}\label{subsec:Environemt}
The experiments were executed on a Cuda (Version 12.4) enabled machine with a GTX 3060 Ti graphics card, and on a high-performance computing cluster. Implementation was done using Python 3.10, as well as the PyTorch and Pandas libraries. 

\subsection{\textbf{Methodology}}\label{subsec:Proccesses}
This section details the methods followed for data preprocessing, hyperparameter tuning, and result generation throughout the experiments.
\subsubsection{\textbf{Data preprocessing}}
For numeric datasets, the following preprocessing steps were taken:
 removing all records with missing features, removing unique identifier features, normalising the numeric features using the $z$-scoring method, and using one-hot encoding to convert categorical features to numeric features.
For the image datasets, the following preprocessing steps were taken:
 resizing the images in the datasets to ensure that all the images in a dataset are the same size, normalising the pixel values across the images, and using one-hot encoding to convert the targets for categorical features to numeric features. {We do not explicitly correct class imbalance to assess the effect of regularisation in realistic settings.}

\subsubsection{\textbf{Training, validation and testing}}
Prior to the hyperparameter tuning process {and training of NN modules}, a testing set was created from each dataset by {using stratified sampling to select} 10\% of the dataset. {Stratified sampling was used in order to ensure class distribution was preserved.} The remaining 90\% of the dataset was used to tune hyperparameters using 15-fold cross-validation.

\subsubsection{\textbf{Hyperparameter tuning}}
The PyHopper library~\cite{lechner2022pyhopper} was used to optimise hyperparameters for each NN model per dataset. PyHopper implements a Markov chain Monte Carlo (MCMC) based optimisation algorithm, and offers a viable alternative to a computationally infeasible grid search.

The PyHopper library was set up to minimise the loss value of a NN over 150 steps, where a step is a single iteration or transition within the Markov chain. The hyperparameters were divided into two groups: (1) basic hyperparameters shared by the baseline NNs (no regularisation) and regularised NNs (e.g., number of layers, learning rate, etc.), and (2) regularisation hyperparameters, which contain the hyperparameters specific to regularisation techniques (e.g., percentage of dropout per layer, pruning amount, etc.). For a full list of hyperparameters and their values, see Tables \ref{tab:NNHyperparameters} and \ref{tab:CNNHyperparameters}. The regularisation hyperparameters were further grouped by regularisation techniques, and tuned individually of each other.
\begin{table*}[]
    \centering
    \caption{Optimised NN hyperparameter values for numeric datasets. ``G'' refers to the hyperparameter optimisation group.}
    \label{tab:NNHyperparameters}
    \begin{tabular}{|l|l|lllll|}
        \hline
        \multirow{2}{*}{\textbf{Hyperparameters}}                                               & \multirow{2}{*}{\textbf{G}} & \multicolumn{5}{c|}{\textbf{Datasets}}                                                                                                                                                                                                                                                                                                                                                                                                     \\ \cline{3-7} 
                                                                                               &                                 & \multicolumn{1}{l|}{\textbf{Diabetes}}                                                                     & \multicolumn{1}{l|}{\textbf{\begin{tabular}[c]{@{}l@{}}Liver Cirrhosis\end{tabular}}}  & \multicolumn{1}{l|}{\textbf{Magic}}     & \multicolumn{1}{l|}{\textbf{Mfeat-pixel}}                                                        & \textbf{\begin{tabular}[c]{@{}l@{}}White Wine Quality\end{tabular}}               \\ \hline
        \textbf{Batch size}                                                                    & 1                               & \multicolumn{1}{l|}{58}                                                                                    & \multicolumn{1}{l|}{1024}                                                                 & \multicolumn{1}{l|}{56}                 & \multicolumn{1}{l|}{42}                                                                          & 32                                                                                   \\ \hline
        \textbf{Dropout layers}                                                                & 2                               & \multicolumn{1}{l|}{\begin{tabular}[c]{@{}l@{}}{[}0.183, 0.006, 0.589, \\0.395, 0.630{]}\end{tabular}} & \multicolumn{1}{l|}{\begin{tabular}[c]{@{}l@{}}{[}0.171, 0.243,\\0.505{]}\end{tabular}} & \multicolumn{1}{l|}{{[}0.346, 0.166{]}} & \multicolumn{1}{l|}{\begin{tabular}[c]{@{}l@{}}{[}0.378, 0.514, 0.288,\\0.733{]}\end{tabular}} & \begin{tabular}[c]{@{}l@{}}{[}0.000, 0.00, 0.012,\\0.579, 0.565{]}\end{tabular} \\ \hline
        \textbf{Learning rate}                                                                 & 1                               & \multicolumn{1}{l|}{0.024}                                                                                 & \multicolumn{1}{l|}{0.593}                                                                & \multicolumn{1}{l|}{0.052}              & \multicolumn{1}{l|}{0.030}                                                                       & 0.008                                                                                \\ \hline
        \textbf{Momentum}                                                                      & 1                               & \multicolumn{1}{l|}{0.033}                                                                                 & \multicolumn{1}{l|}{0.900}                                                                & \multicolumn{1}{l|}{0.018}              & \multicolumn{1}{l|}{0.071}                                                                       & 0.058                                                                                \\ \hline
        \textbf{No. of epochs}                                                              & 1                               & \multicolumn{1}{l|}{190}                                                                                   & \multicolumn{1}{l|}{500}                                                                  & \multicolumn{1}{l|}{100}                & \multicolumn{1}{l|}{500}                                                                         & 220                                                                                  \\ \hline
        \textbf{\begin{tabular}[c]{@{}l@{}}No. of hidden layers\end{tabular}}            & 1                               & \multicolumn{1}{l|}{5}                                                                                     & \multicolumn{1}{l|}{3}                                                                    & \multicolumn{1}{l|}{2}                  & \multicolumn{1}{l|}{4}                                                                           & 5                                                                                    \\ \hline
        \textbf{\begin{tabular}[c]{@{}l@{}}No. of neurons in layers\end{tabular}}        & 1                               & \multicolumn{1}{l|}{\begin{tabular}[c]{@{}l@{}}{[}150, 60, 200, 30, 20{]}\end{tabular}}                 & \multicolumn{1}{l|}{\begin{tabular}[c]{@{}l@{}}{[}800, 850, 100{]}\end{tabular}}       & \multicolumn{1}{l|}{{[}250, 90{]}}      & \multicolumn{1}{l|}{\begin{tabular}[c]{@{}l@{}}{[}180, 120, 70, 160{]}\end{tabular}}          & \begin{tabular}[c]{@{}l@{}}{[}300, 250, 200, 850,\\400{]}\end{tabular}             \\ \hline
        \textbf{Prune amount}                                                                  & \multirow{2}{*}{2}              & \multicolumn{1}{l|}{0.266}                                                                                 & \multicolumn{1}{l|}{0.312}                                                                & \multicolumn{1}{l|}{0.302}              & \multicolumn{1}{l|}{0.199}                                                                       & 0.76                                                                                 \\ \cline{1-1} \cline{3-7} 
        \textbf{Prune epoch interval}                                                          &                                 & \multicolumn{1}{l|}{19}                                                                                    & \multicolumn{1}{l|}{50}                                                                   & \multicolumn{1}{l|}{10}                 & \multicolumn{1}{l|}{50}                                                                          & 38                                                                                   \\ \hline
        \textbf{Weight decay}                                                                  & 2                               & \multicolumn{1}{l|}{0.001}                                                                                 & \multicolumn{1}{l|}{0.001}                                                                & \multicolumn{1}{l|}{0.002}              & \multicolumn{1}{l|}{0.002}                                                                       & 0.001                                                                                \\ \hline
        \textbf{\begin{tabular}[c]{@{}l@{}}Weight perturb. amount\end{tabular}}         & \multirow{2}{*}{2}              & \multicolumn{1}{l|}{0.010}                                                                                 & \multicolumn{1}{l|}{0.044}                                                                & \multicolumn{1}{l|}{0.025}              & \multicolumn{1}{l|}{0.033}                                                                       & 0.050                                                                                \\ \cline{1-1} \cline{3-7} 
        \textbf{\begin{tabular}[c]{@{}l@{}}Weight perturb. epoch interval\end{tabular}} &                                 & \multicolumn{1}{l|}{22}                                                                                    & \multicolumn{1}{l|}{50}                                                                   & \multicolumn{1}{l|}{17}                 & \multicolumn{1}{l|}{5}                                                                           & 33                                                                                   \\ \hline
    \end{tabular}%
\end{table*}

\begin{table*}[]
        \centering
        \caption{Optimised CNN hyperparameter values for image datasets. ``G" refers to the hyperparameter optimisation group.}
        \label{tab:CNNHyperparameters}
    \begin{tabular}{|l|l|lllll|}
    \hline
    \multirow{2}{*}{\textbf{Hyperparameters}}                                               & \multirow{2}{*}{\textbf{G}} & \multicolumn{5}{c|}{\textbf{Datasets}}                                                                                                                                                                                                                                                                                                                                                                                                                                         \\ \cline{3-7} 
                                                                                           &                                 & \multicolumn{1}{l|}{\textbf{Balls}}                                                              & \multicolumn{1}{l|}{\textbf{Bean Leafs}}                                                                   & \multicolumn{1}{l|}{\textbf{Cifar 10}}                                                                                             & \multicolumn{1}{l|}{\textbf{Fashion MNIST}} & \textbf{Shoes}                                                              \\ \hline
    \textbf{Batch size}                                                                    & 1                               & \multicolumn{1}{l|}{240}                                                                         & \multicolumn{1}{l|}{128}                                                                                   & \multicolumn{1}{l|}{32}                                                                                                            & \multicolumn{1}{l|}{128}                    & 16                                                                          \\ \hline
    \textbf{Dropout layers} & 2 & \multicolumn{1}{l|}{\begin{tabular}[c]{@{}l@{}}{[}0.370, 0.685, 0.304,\\0.470{]}\end{tabular}} & \multicolumn{1}{l|}{\begin{tabular}[c]{@{}l@{}}{[}0.015, 0.005, 0.493,\\0.130, 0.377{]}\end{tabular}} & \multicolumn{1}{l|}{\begin{tabular}[c]{@{}l@{}}{[}0.300, 0.300, 0.000,\\0.000, 0.300, 0.500,\\ 0.500, 0.500{]}\end{tabular}} & \multicolumn{1}{l|}{\begin{tabular}[c]{@{}l@{}}{[}0.000, 0.250, 0.000,\\0.250, 0.000, 0.250,\\0.000, 0.250, 0.000,\\0.250, 0.250, 0.250{]}\end{tabular}} & \multicolumn{1}{l|}{\begin{tabular}[c]{@{}l@{}}{[}0.005, 0.006,\\0.791, 0.057{]}\end{tabular}} \\ \hline    
    \textbf{Kernel size}                                                                   & 1                               & \multicolumn{1}{l|}{{[}3x3, 3x3{]}}                                                              & \multicolumn{1}{l|}{{[}2x2, 2x2{]}}                                                                        & \multicolumn{1}{l|}{\begin{tabular}[c]{@{}l@{}}{[}3x3, 3x3, 3x3, 3x3,\\3x3{]}\end{tabular}}                                   & \multicolumn{1}{l|}{\begin{tabular}[c]{@{}l@{}}{[}3x3, 3x3, 3x3, 3x3,\\2x2{]}\end{tabular}}                        & {[}16x16, 2x2{]}                                                            \\ \hline
    \textbf{Kernel stride}                                                                 & 1                               & \multicolumn{1}{l|}{{[}3, 3{]}}                                                                  & \multicolumn{1}{l|}{{[}2, 2{]}}                                                                            & \multicolumn{1}{l|}{{[}1, 1, 1, 1, 1{]}}                                                                                           & \multicolumn{1}{l|}{{[}1, 1, 1, 1, 1{]}}                       & {[}4, 2{]}                                                                  \\ \hline
    \textbf{Learning rate}                                                                 & 1                               & \multicolumn{1}{l|}{0.250}                                                                       & \multicolumn{1}{l|}{0.250}                                                                                 & \multicolumn{1}{l|}{0.002}                                                                                                         & \multicolumn{1}{l|}{0.016}                       & 0.026                                                                       \\ \hline
    \textbf{Momentum}                                                                      & 1                               & \multicolumn{1}{l|}{0.009}                                                                       & \multicolumn{1}{l|}{0.001}                                                                                 & \multicolumn{1}{l|}{0.098}                                                                                                         & \multicolumn{1}{l|}{0.075}                       & 0.027                                                                       \\ \hline
    \textbf{\begin{tabular}[c]{@{}l@{}}No. of convolutional layers\end{tabular}}     & 1                               & \multicolumn{1}{l|}{2}                                                                           & \multicolumn{1}{l|}{2}                                                                                     & \multicolumn{1}{l|}{5}                                                                                                             & \multicolumn{1}{l|}{5}                       & 2                                                                           \\ \hline
    \textbf{No. of epochs}                                                              & 1                               & \multicolumn{1}{l|}{500}                                                                         & \multicolumn{1}{l|}{200}                                                                                   & \multicolumn{1}{l|}{200}                                                                                                           & \multicolumn{1}{l|}{100}                       & 300                                                                         \\ \hline
    \textbf{\begin{tabular}[c]{@{}l@{}}No. of hidden layers\end{tabular}}            & 1                               & \multicolumn{1}{l|}{2}                                                                           & \multicolumn{1}{l|}{3}                                                                                     & \multicolumn{1}{l|}{3}                                                                                                             & \multicolumn{1}{l|}{2}                       & 2                                                                           \\ \hline
    \textbf{\begin{tabular}[c]{@{}l@{}}No. of neurons in layers\end{tabular}}        & 1                               & \multicolumn{1}{l|}{{[}50, 50{]}}                                                                & \multicolumn{1}{l|}{{[}475, 450, 425{]}}                                                                   & \multicolumn{1}{l|}{{[}512, 256, 128{]}}                                                                                           & \multicolumn{1}{l|}{{[}512, 64{]}}                       & {[}375, 300{]}                                                              \\ \hline
    \textbf{Out channels}                                                                  & 1                               & \multicolumn{1}{l|}{{[}32, 32{]}}                                                                & \multicolumn{1}{l|}{{[}4, 16{]}}                                                                           & \multicolumn{1}{l|}{\begin{tabular}[c]{@{}l@{}}{[}128, 256, 512, 512,\\256{]}\end{tabular}}                                      & \multicolumn{1}{l|}{\begin{tabular}[c]{@{}l@{}}{[}64, 64, 128, 128,\\256{]}\end{tabular}}                        & {[}64, 8{]}                                                                 \\ \hline
    \textbf{Padding}                                                                       & 1                               & \multicolumn{1}{l|}{{[}0, 0{]}}                                                                  & \multicolumn{1}{l|}{{[}0, 0{]}}                                                                            & \multicolumn{1}{l|}{{[}1, 1, 1, 1, 1{]}}                                                                                           & \multicolumn{1}{l|}{{[}1, 1, 1, 1, 1{]}}                       & {[}0,0{]}                                                                   \\ \hline
    \textbf{Pool size}                                                                     & 1                               & \multicolumn{1}{l|}{{[}2, 2{]}}                                                                  & \multicolumn{1}{l|}{{[}4, 2{]}}                                                                            & \multicolumn{1}{l|}{{[}2, 2, 2, 2, 2{]}}                                                                                           & \multicolumn{1}{l|}{{[}2, 2, 2, 2, 2{]}}                       & {[}4, 2{]}                                                                  \\ \hline
    \textbf{Pool type}                                                                     & 1                               & \multicolumn{1}{l|}{{[}max, max{]}}                                                              & \multicolumn{1}{l|}{{[}avg, avg{]}}                                                                      & \multicolumn{1}{l|}{\begin{tabular}[c]{@{}l@{}}{[}max, max, max, avg,\\avg{]}\end{tabular}}                                    & \multicolumn{1}{l|}{\begin{tabular}[c]{@{}l@{}}{[}avg, avg, max, avg,\\max{]}\end{tabular}}                       & {[}max, avg{]}                                                             \\ \hline
    \textbf{Prune amount}                                                                  & \multirow{2}{*}{2}              & \multicolumn{1}{l|}{0.453}                                                                       & \multicolumn{1}{l|}{0.361}                                                                                 & \multicolumn{1}{l|}{0.427}                                                                                                         & \multicolumn{1}{l|}{0.005}                       & 0.564                                                                       \\ \cline{1-1} \cline{3-7} 
    \textbf{Prune epoch interval}                                                          &                                 & \multicolumn{1}{l|}{75}                                                                          & \multicolumn{1}{l|}{30}                                                                                    & \multicolumn{1}{l|}{75}                                                                                                            & \multicolumn{1}{l|}{0.030}                       & 65                                                                          \\ \hline
    \textbf{Weight decay}                                                                  & 2                               & \multicolumn{1}{l|}{0.005}                                                                       & \multicolumn{1}{l|}{0.001}                                                                                 & \multicolumn{1}{l|}{0.001}                                                                                                         & \multicolumn{1}{l|}{10}                       & 0.010                                                                 \\ \hline
    \textbf{\begin{tabular}[c]{@{}l@{}}Weight perturbation amount\end{tabular}}         & \multirow{2}{*}{2}              & \multicolumn{1}{l|}{0.100}                                                                       & \multicolumn{1}{l|}{0.250}                                                                                 & \multicolumn{1}{l|}{0.010}                                                                                                         & \multicolumn{1}{l|}{0.050}                       & 0.05                                                                        \\ \cline{1-1} \cline{3-7} 
    \textbf{\begin{tabular}[c]{@{}l@{}}Weight perturbation \\ epoch interval\end{tabular}} &                                 & \multicolumn{1}{l|}{25}                                                                          & \multicolumn{1}{l|}{25}                                                                                    & \multicolumn{1}{l|}{5}                                                                                                             & \multicolumn{1}{l|}{10}                       & 10                                                                          \\ \hline
    \end{tabular}%
\end{table*}

\subsubsection{\textbf{Result generation}}
For each regularisation technique, as well as the non-regularised baseline, each NN was trained 15 independent times. For each epoch of each independent run, the testing $F1$ scores were recorded, together with the random starting seeds. The testing $F1$ scores are used to illustrate and compare the impact of the regularisation techniques on the generalisation ability of NN.

\subsection{\textbf{Results and Discussion}}\label{subsec:results}
\begin{figure*}[htbp]
    \centering
    \begin{subfigure}{1\columnwidth}
        \centering
        \includegraphics[width=1\columnwidth]{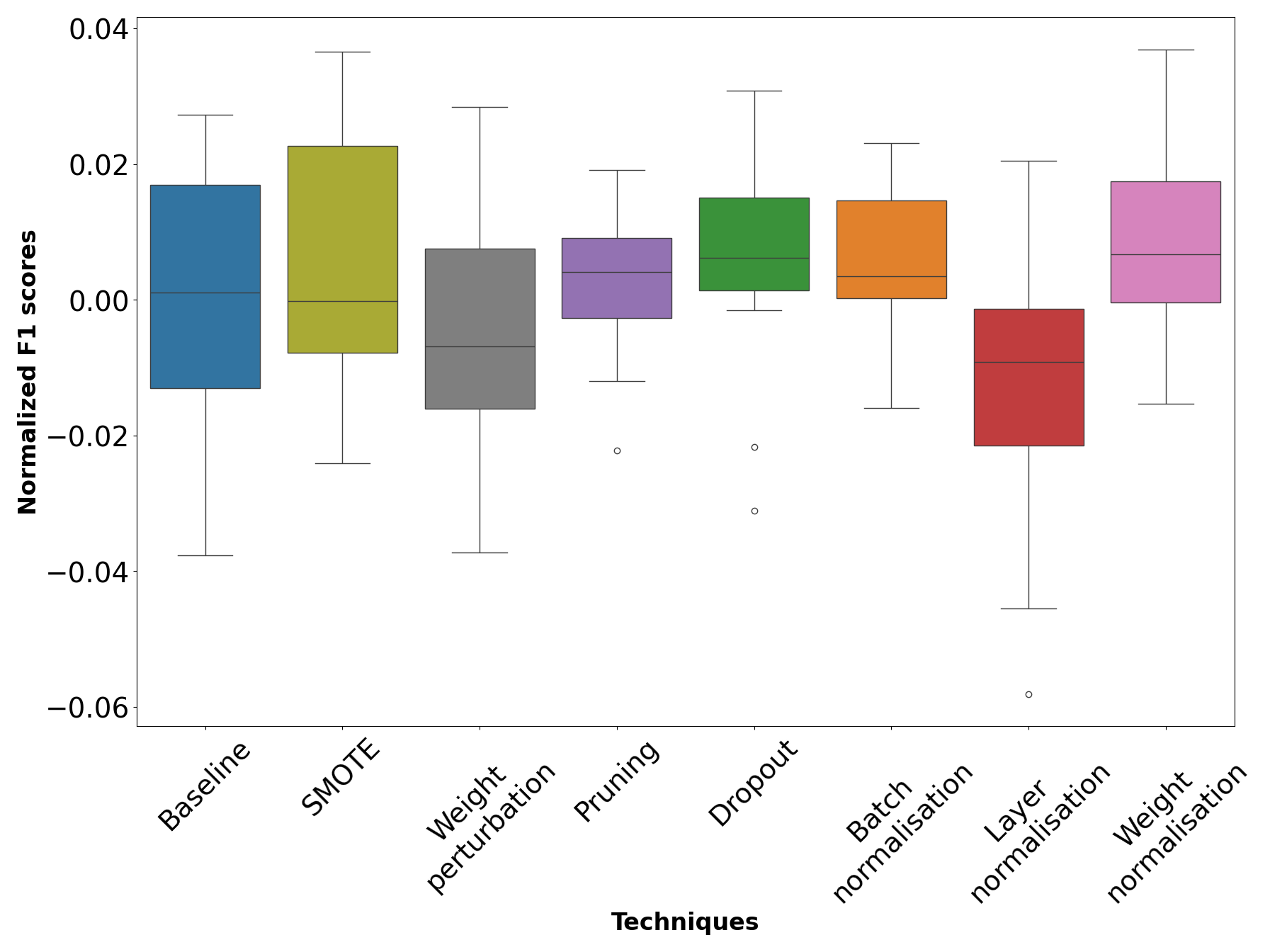}
        \caption{Diabetes Dataset}
    \end{subfigure}  
    \begin{subfigure}{1\columnwidth}
        \centering
        \includegraphics[width=1\columnwidth]{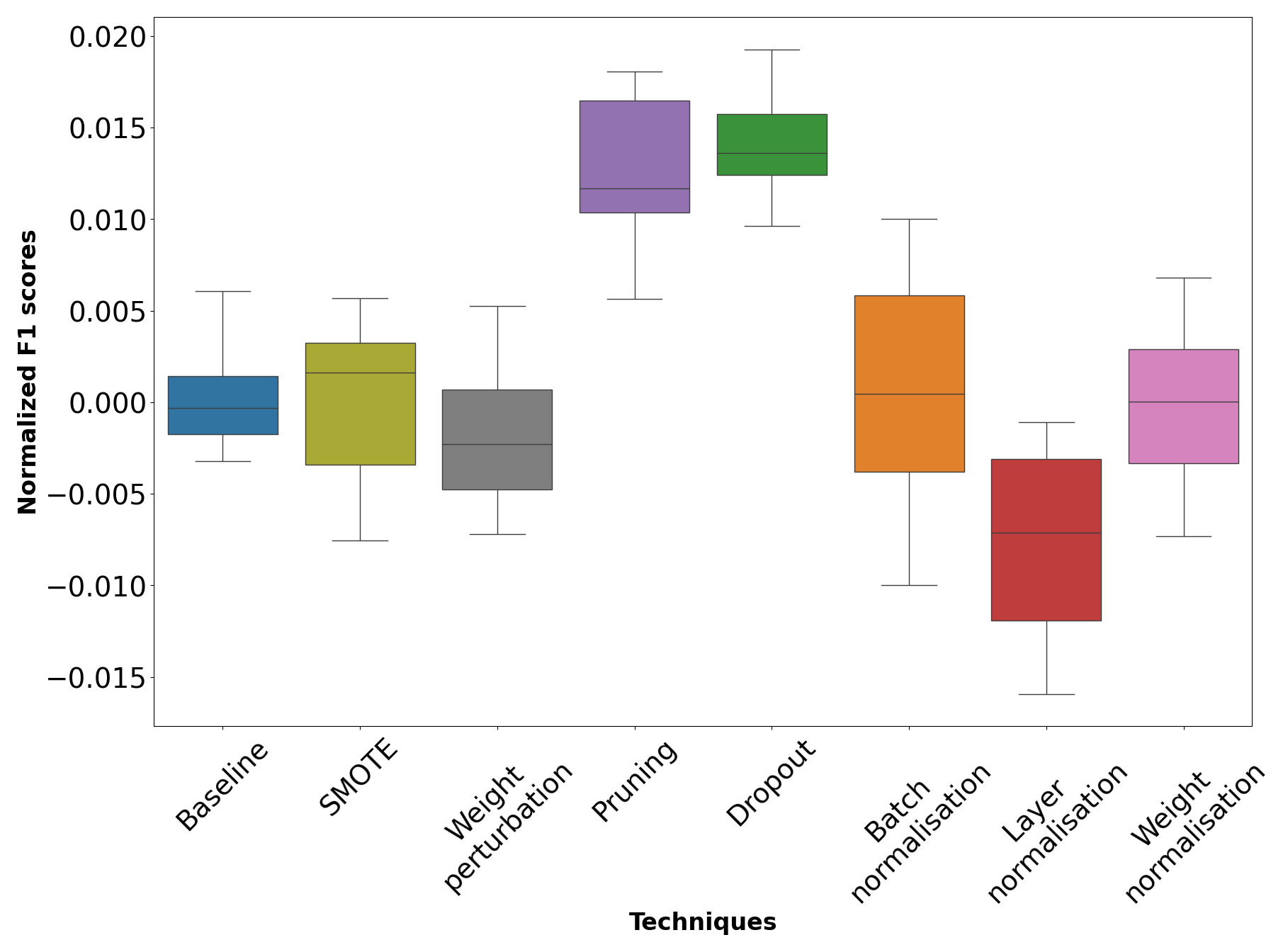}
        \caption{Liver cirrhosis Dataset}
    \end{subfigure}
    \begin{subfigure}{1\columnwidth}
        \centering
        \includegraphics[width=1\columnwidth]{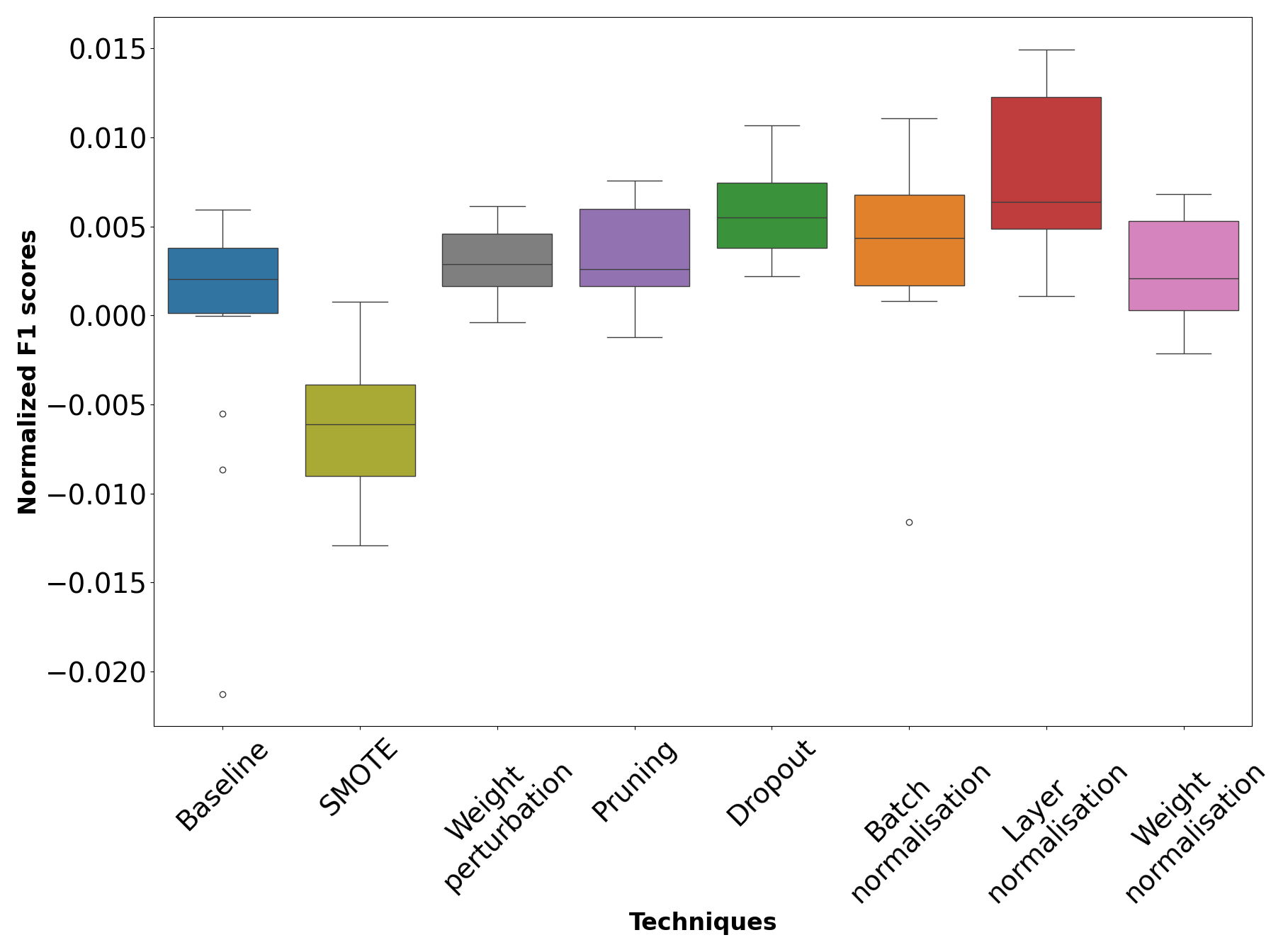}
        \caption{Magic Dataset}
    \end{subfigure}
    \begin{subfigure}{1\columnwidth}
        \centering
        \includegraphics[width=1\columnwidth]{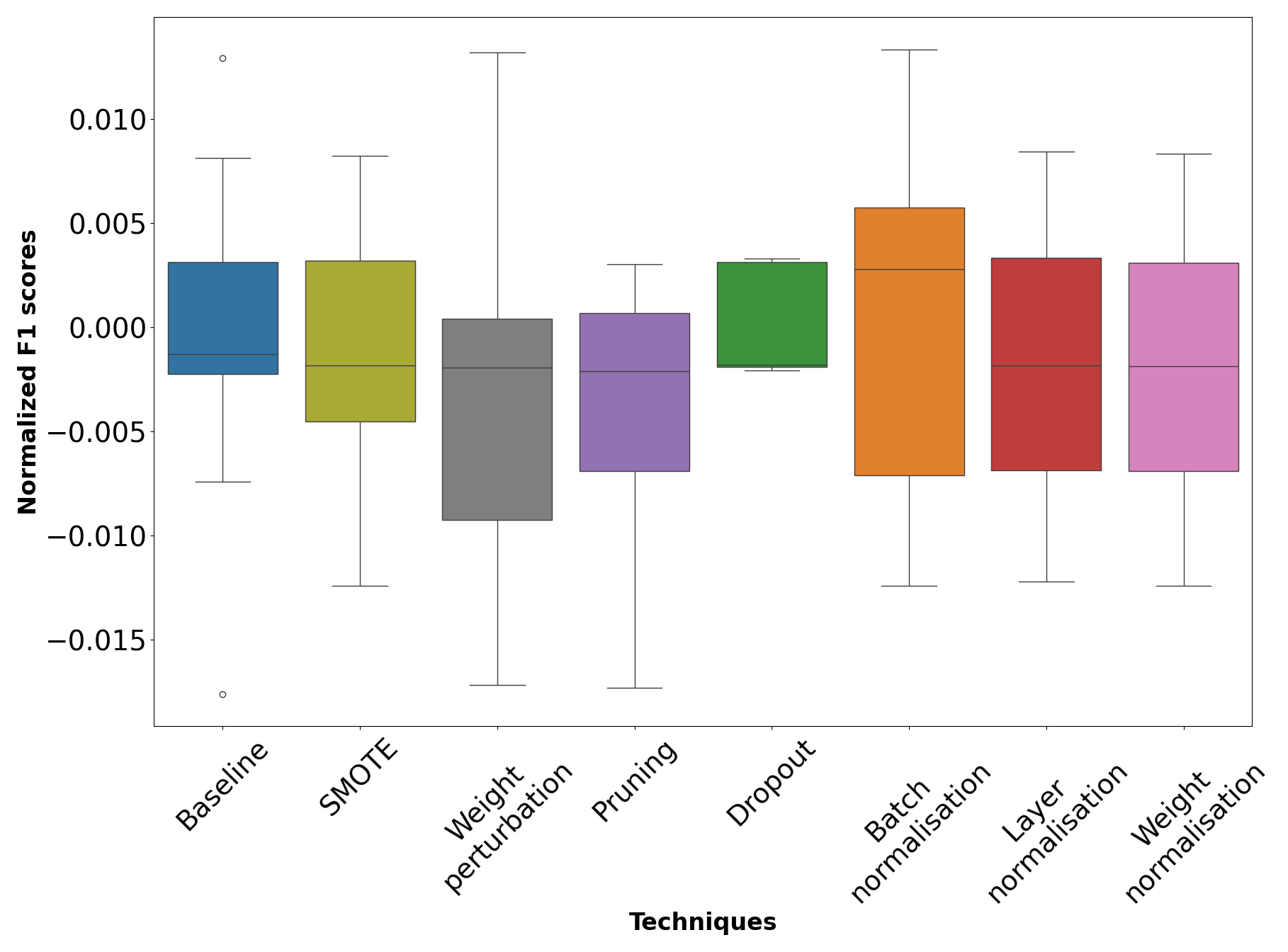}
        \caption{Mfeat pixel Dataset}
    \end{subfigure}
    \begin{subfigure}{1\columnwidth}
        \centering
        \includegraphics[width=1\columnwidth]{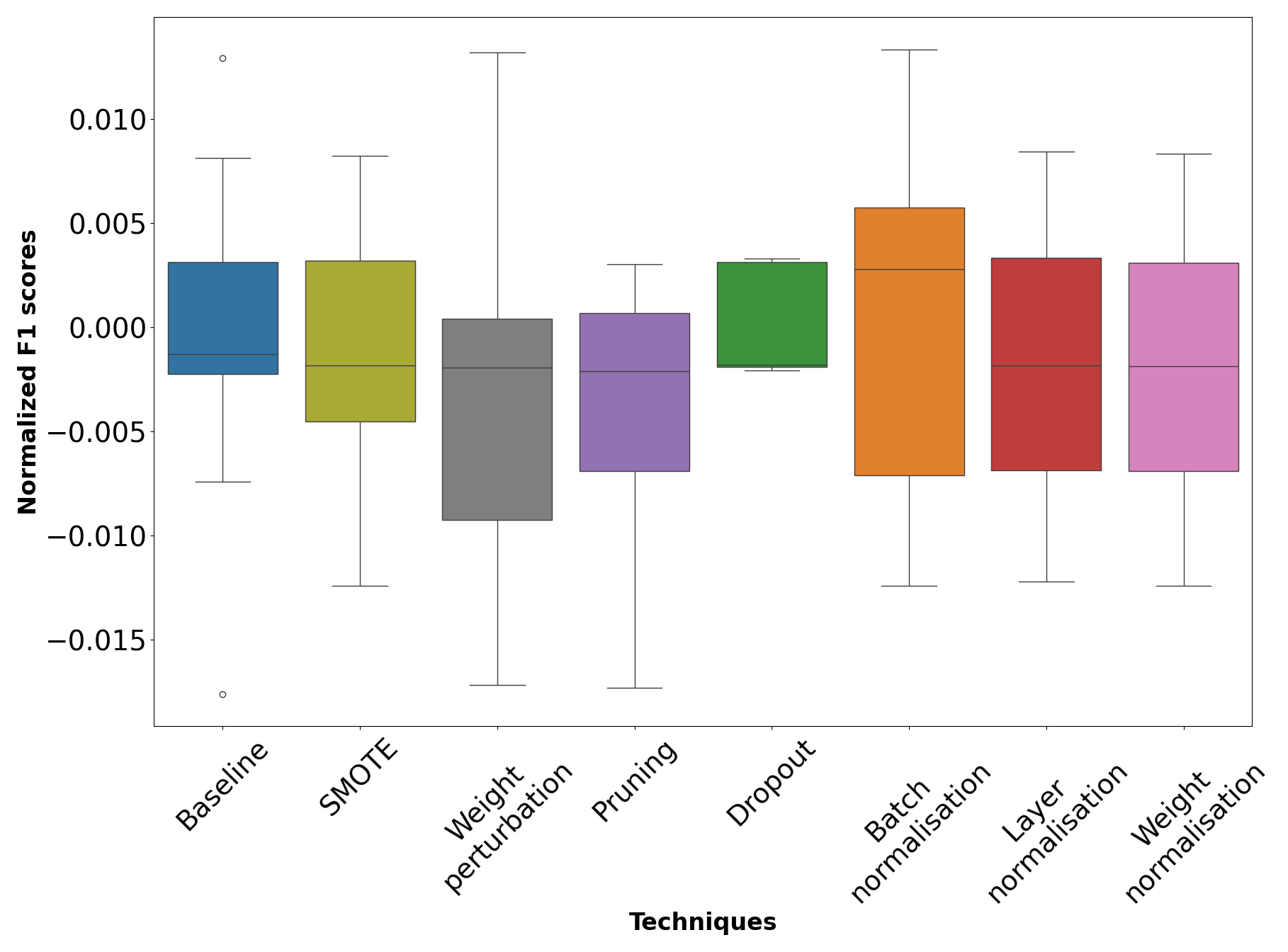}
        \caption{White wine quality Dataset}
    \end{subfigure}
    
    \caption{Box and whisker charts of the normalized testing F1 scores for numeric datasets.}
    \label{fig:NumericCharts}
\end{figure*}

\begin{figure*}[htbp]
    \centering
    \begin{subfigure}{1\columnwidth}
        \centering
        \includegraphics[width=1\columnwidth]{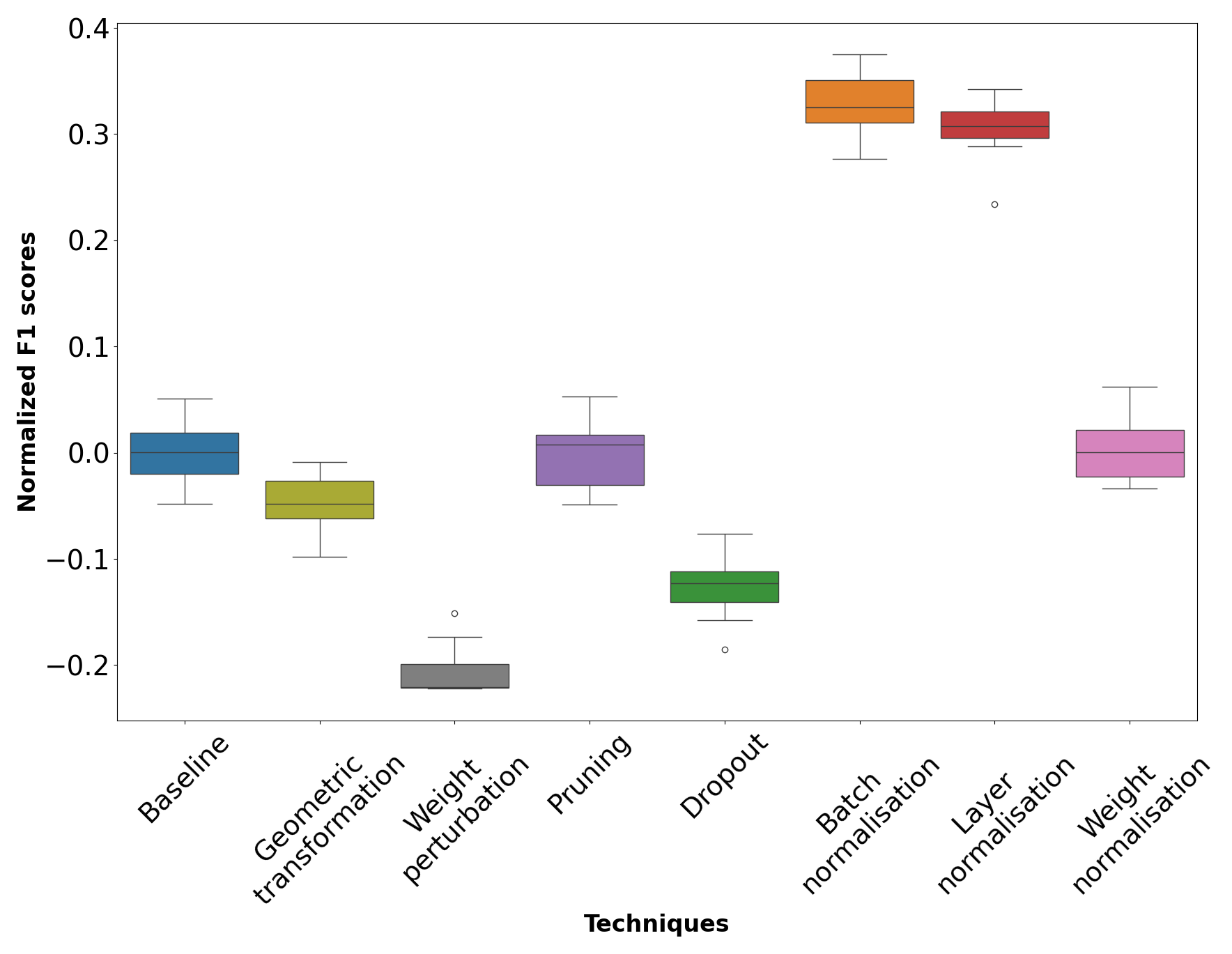}
        \caption{Balls Dataset}
    \end{subfigure}  
    \hfill 
    \begin{subfigure}{1\columnwidth}
        \centering
        \includegraphics[width=1\columnwidth]{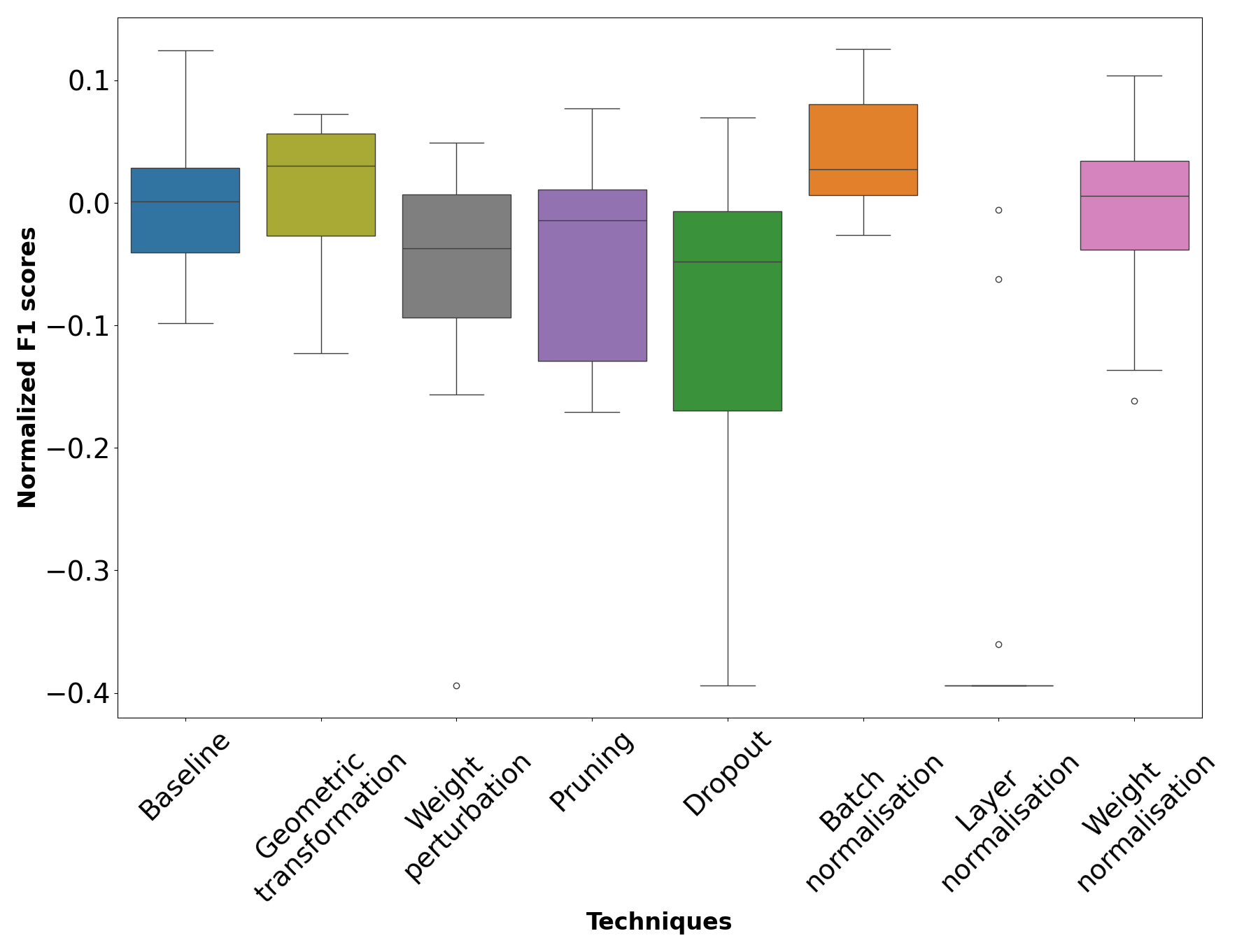}
        \caption{Bean Leafs Dataset}
    \end{subfigure}
    \hfill 
    \begin{subfigure}{1\columnwidth}
        \centering
        \includegraphics[width=1\columnwidth]{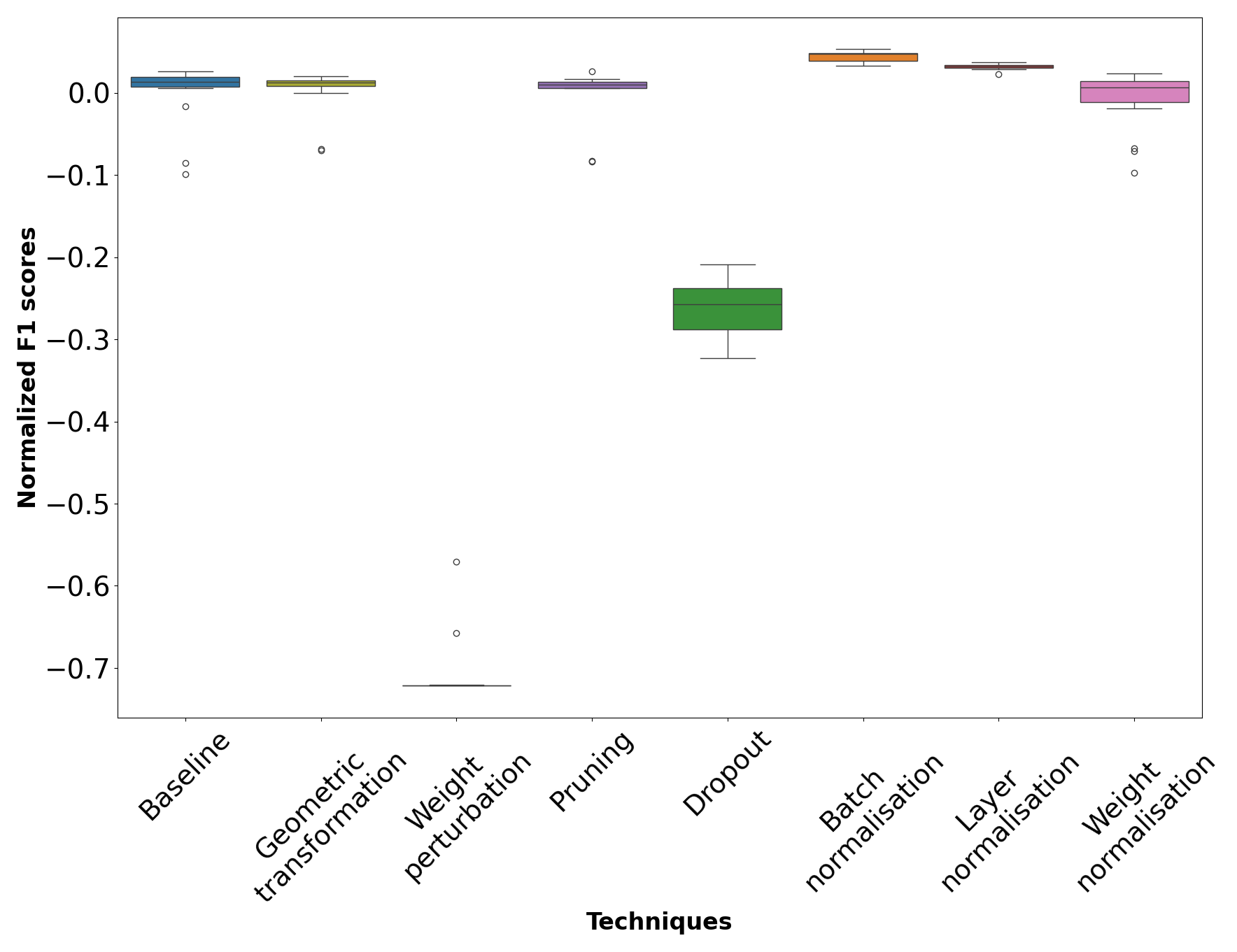}
        \caption{Cifar10 Dataset}
    \end{subfigure}
    \hfill 
    \begin{subfigure}{1\columnwidth}
        \centering
        \includegraphics[width=1\columnwidth]{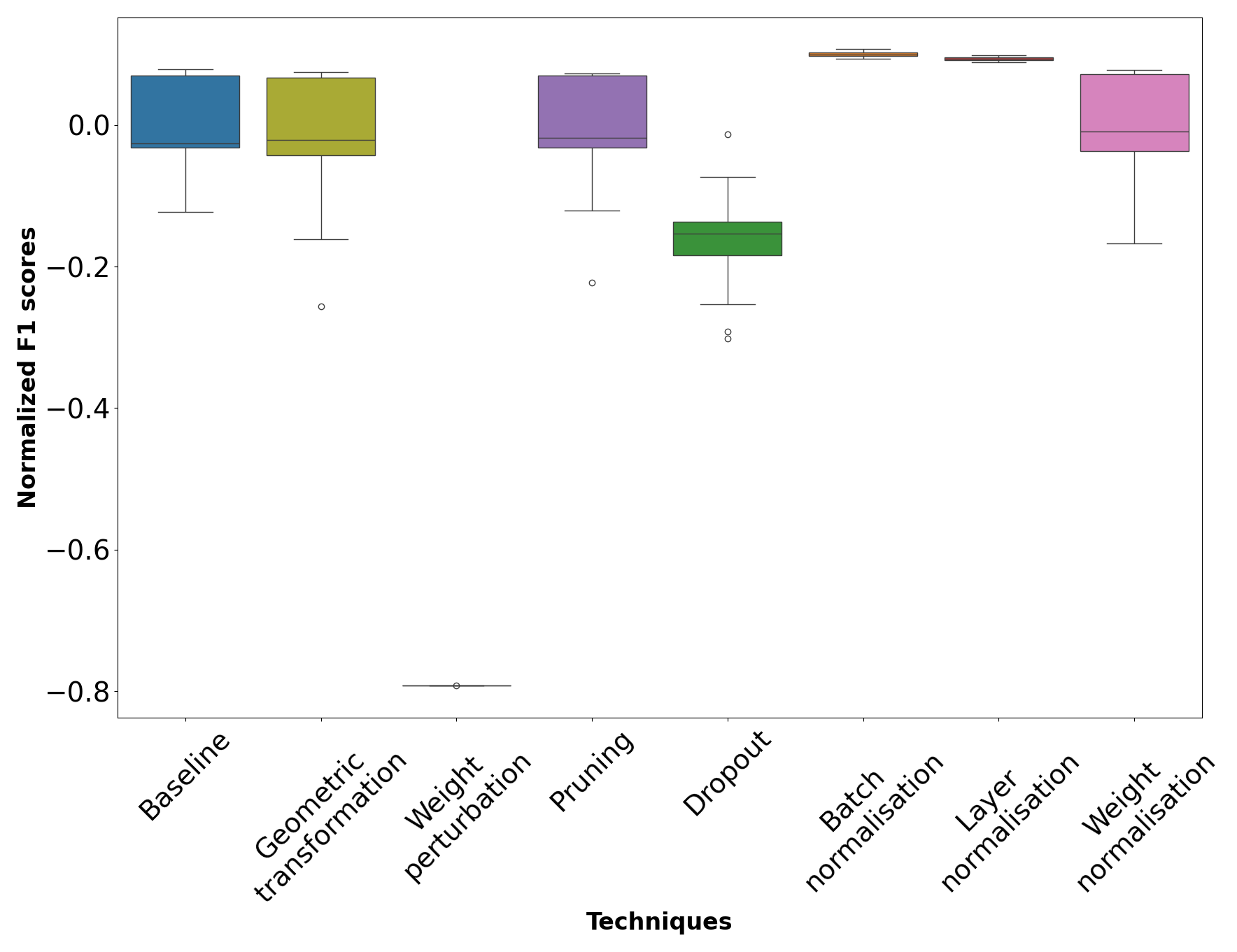}
        \caption{Fashion MNIST Dataset}
    \end{subfigure}
    \hfill 
    \begin{subfigure}{1\columnwidth}
        \centering
        \includegraphics[width=1\columnwidth]{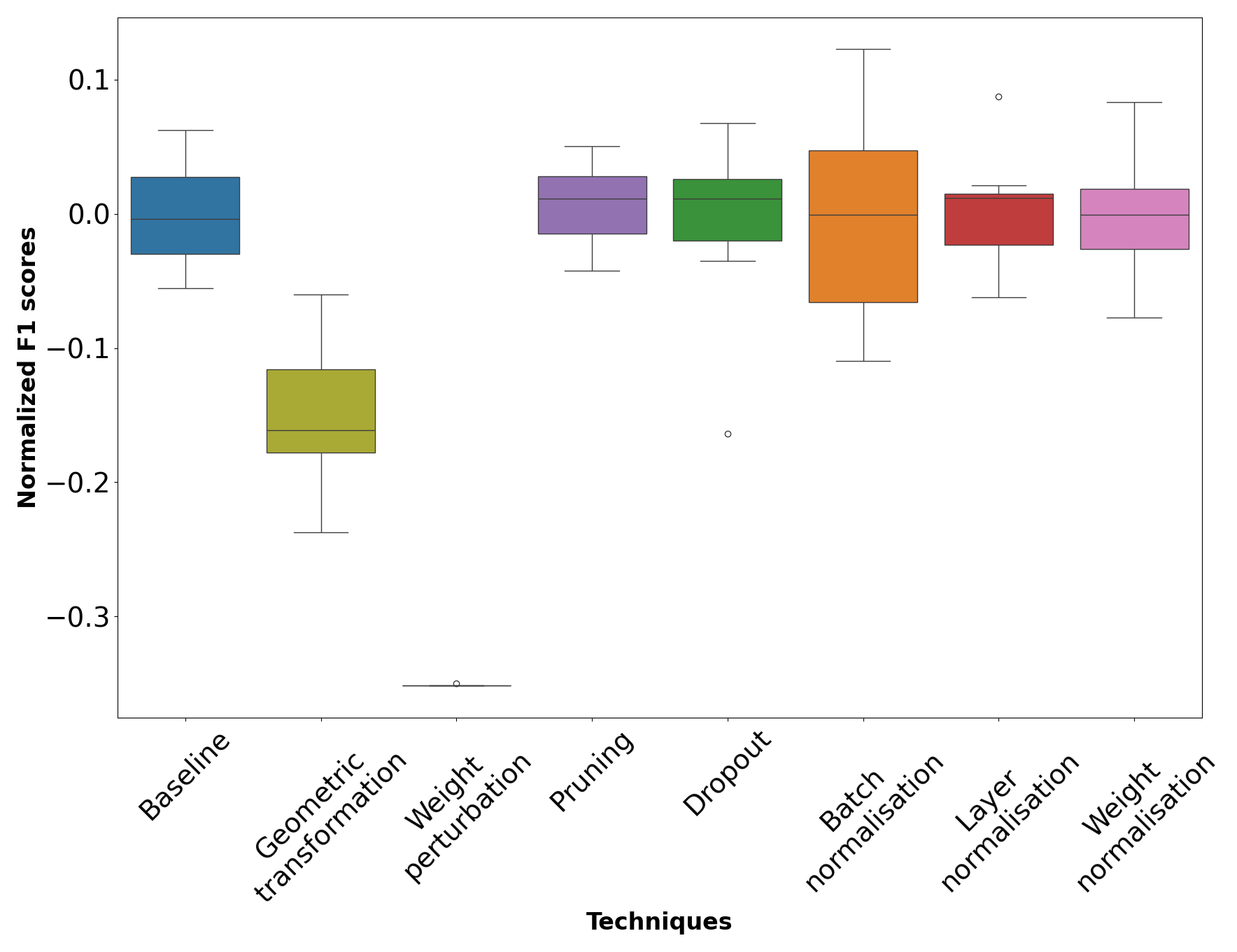}
        \caption{Shoes Dataset}
    \end{subfigure}
    
    \caption{Box and whisker charts of the testing F1 scores for various image datasets.}
    \label{fig:ImagesCharts}
\end{figure*}

The results are summarised in Table~\ref{tab:AvgAndStdOfF1Scores} and illustrated as box and whiskers plots in Fig.~\ref{fig:NumericCharts} and Fig.~\ref{fig:ImagesCharts}. To aid the comparability of the results, all values in the box plots were normalised by calculating the mean of the baseline experiments (the 15 runs where no regularisation techniques were used), and subtracting this mean from each result. As such, negative normalised $F1$-score values correspond to performance that is worse than the baseline. For the values in Table~\ref{tab:AvgAndStdOfF1Scores}, Mann-Whitney U hypothesis testing was performed (significance level of $0.005$), with the null hypothesis being that the performance of a NN with a specific regularisation technique does not statistically differ from the baseline. {In Table~\ref{tab:AvgAndStdOfF1Scores},} the values that are statistically better than the baseline are indicated in bold and blue, and values that are statistically worse than the baseline are indicated in bold and orange. 

\begin{table*}[]
    \centering
    \caption{The mean and standard deviation of the testing F1 scores for datasets. Highlighted values in \textcolor[HTML]{DF6747}{\textbf{orange}} indicate significant decrease, while those in \textcolor[HTML]{0F2080}{\textbf{blue}} indicate significant improvement over the baseline.}
    \label{tab:AvgAndStdOfF1Scores}
    \begin{tabular}{|l|l|c|c|c|c|c|c|c|c|c|c|}
		\hline
		\multicolumn{1}{|c|}{\multirow{2}{*}{}}                                     & \multirow{2}{*}{} 						& \multicolumn{10}{c|}{\textbf{Mean F1 $\pm$ standard deviation}} \\ \cline{3-12} 
		\multicolumn{1}{|c|}{}                                                      &					 						& \multicolumn{5}{c|}{\textbf{Image datasets}}																														& \multicolumn{5}{c|}{\textbf{Numeric Datasets}}\\ \hline
		\multicolumn{1}{|c|}{\textbf{Name}}											& \multicolumn{1}{l|}{\textbf{{\begin{tabular}[c]{@{}l@{}}Regularisation\\strategy\end{tabular}}}} 	& \textbf{Balls}																																					& \textbf{\begin{tabular}[c]{@{}l@{}}Bean\\leafs\end{tabular}}& \textbf{Cifar 10}& \textbf{\begin{tabular}[c]{@{}l@{}}Fashion\\MNIST\end{tabular}}& \textbf{Shoes}& \textbf{Diabetes}& \textbf{\begin{tabular}[c]{@{}l@{}}Liver\\cirrhosis\end{tabular}}& \textbf{Magic}& \textbf{\begin{tabular}[c]{@{}l@{}}Mfeat-\\pixel\end{tabular}}& \textbf{\begin{tabular}[c]{@{}l@{}}Wine\\quality\end{tabular}}\\ \hline
		\textbf{Baseline}															& \multicolumn{1}{l|}{{n.a.}}					& \multicolumn{1}{r|}{\begin{tabular}[c]{@{}r@{}}0.224\\$\pm$0.027\end{tabular}}& \multicolumn{1}{r|}{\begin{tabular}[c]{@{}r@{}}0.561\\$\pm$0.052\end{tabular}}	& \multicolumn{1}{r|}{\begin{tabular}[c]{@{}r@{}}0.740\\$\pm$0.038\end{tabular}}& \multicolumn{1}{r|}{\begin{tabular}[c]{@{}r@{}}0.810\\$\pm$0.072\end{tabular}}& \multicolumn{1}{r|}{\begin{tabular}[c]{@{}r@{}}0.518\\$\pm$0.038\end{tabular}}& \multicolumn{1}{r|}{\begin{tabular}[c]{@{}r@{}}0.733\\$\pm$0.019\end{tabular}}	& \multicolumn{1}{r|}{\begin{tabular}[c]{@{}r@{}}0.911\\$\pm$0.003\end{tabular}}& \multicolumn{1}{r|}{\begin{tabular}[c]{@{}r@{}}0.861\\$\pm$0.007\end{tabular}}& \multicolumn{1}{r|}{\begin{tabular}[c]{@{}r@{}}0.967\\$\pm$0.007\end{tabular}}	& \multicolumn{1}{r|}{\begin{tabular}[c]{@{}r@{}}0.462\\$\pm$0.006\end{tabular}}\\ \hline\hline
		\textbf{\begin{tabular}[c]{@{}l@{}}Geometric\\transformation\end{tabular}}	& \multicolumn{1}{l|}{{\begin{tabular}[c]{@{}l@{}}Data-based\end{tabular}}}      & \multicolumn{1}{r|}{{\color[HTML]{DF6747} \textbf{\begin{tabular}[c]{@{}r@{}}0.176\\$\pm$0.027\end{tabular}}}}													& \multicolumn{1}{r|}{\begin{tabular}[c]{@{}r@{}}0.564\\$\pm$0.066\end{tabular}}& \multicolumn{1}{r|}{\begin{tabular}[c]{@{}r@{}}0.741\\$\pm$0.028\end{tabular}}& \multicolumn{1}{r|}{\begin{tabular}[c]{@{}r@{}}0.793\\$\pm$0.100\end{tabular}}& \multicolumn{1}{r|}{{\color[HTML]{DF6747} \textbf{\begin{tabular}[c]{@{}r@{}}0.368\\ $\pm$0.049\end{tabular}}}}& \multicolumn{1}{c|}{n.a.}& \multicolumn{1}{c|}{n.a.}& \multicolumn{1}{c|}{n.a.}& \multicolumn{1}{c|}{n.a.}& \multicolumn{1}{c|}{n.a.}\\ \hline
		\textbf{SMOTE}																& \multicolumn{1}{l|}{{\begin{tabular}[c]{@{}l@{}}Data-based\end{tabular}}}      & \multicolumn{1}{c|}{n.a.}																																			& \multicolumn{1}{c|}{n.a.}& \multicolumn{1}{c|}{n.a.}& \multicolumn{1}{c|}{n.a.}& \multicolumn{1}{c|}{n.a.}& \multicolumn{1}{r|}{\begin{tabular}[c]{@{}r@{}}0.739\\$\pm$0.019\end{tabular}}& \multicolumn{1}{r|}{\begin{tabular}[c]{@{}r@{}}0.911\\$\pm$0.004\end{tabular}}& \multicolumn{1}{r|}{{\color[HTML]{DF6747} \textbf{\begin{tabular}[c]{@{}r@{}}0.855\\ $\pm$0.004\end{tabular}}}}	& \multicolumn{1}{r|}{\begin{tabular}[c]{@{}r@{}}0.966\\$\pm$0.007\end{tabular}}	& \multicolumn{1}{c|}{n.a.}\\ \hline
		\textbf{\begin{tabular}[c]{@{}l@{}}Weight\\perturbation\end{tabular}}		& \multicolumn{1}{l|}{{\begin{tabular}[c]{@{}l@{}}Data-based\end{tabular}}}      & \multicolumn{1}{r|}{{\color[HTML]{DF6747} \textbf{\begin{tabular}[c]{@{}r@{}}0.017\\$\pm$0.021\end{tabular}}}}													& \multicolumn{1}{r|}{\begin{tabular}[c]{@{}r@{}}0.501\\$\pm$0.109\end{tabular}}& \multicolumn{1}{r|}{{\color[HTML]{DF6747} \textbf{\begin{tabular}[c]{@{}r@{}}0.033\\$\pm$0.040\end{tabular}}}}& \multicolumn{1}{r|}{{\color[HTML]{DF6747} \textbf{\begin{tabular}[c]{@{}r@{}}0.018\\$\pm$0.000\end{tabular}}}}	& \multicolumn{1}{r|}{{\color[HTML]{DF6747} \textbf{\begin{tabular}[c]{@{}r@{}}0.167\\$\pm$0.000\end{tabular}}}}	& \multicolumn{1}{r|}{\begin{tabular}[c]{@{}r@{}}0.728\\$\pm$0.018\end{tabular}}	& \multicolumn{1}{r|}{\begin{tabular}[c]{@{}r@{}}0.909\\$\pm$0.004\end{tabular}}									& \multicolumn{1}{r|}{\begin{tabular}[c]{@{}r@{}}0.864\\$\pm$0.002\end{tabular}}									& \multicolumn{1}{r|}{\begin{tabular}[c]{@{}r@{}}0.963\\$\pm$0.008\end{tabular}}	& \multicolumn{1}{r|}{\begin{tabular}[c]{@{}r@{}}0.463\\$\pm$0.006\end{tabular}}									\\ \hline\hline
		\textbf{Pruning} 															& \multicolumn{1}{l|}{{\begin{tabular}[c]{@{}l@{}}Architecture\end{tabular}}}    & \multicolumn{1}{r|}{\begin{tabular}[c]{@{}r@{}}0.221\\$\pm$0.03\end{tabular}}																						& \multicolumn{1}{r|}{\begin{tabular}[c]{@{}r@{}}0.515\\$\pm$0.081\end{tabular}}									& \multicolumn{1}{r|}{\begin{tabular}[c]{@{}r@{}}0.739\\$\pm$0.033\end{tabular}}									& \multicolumn{1}{r|}{\begin{tabular}[c]{@{}r@{}}0.803\\$\pm$0.082\end{tabular}}									& \multicolumn{1}{r|}{\begin{tabular}[c]{@{}r@{}}0.527\\$\pm$0.028\end{tabular}}									& \multicolumn{1}{r|}{\begin{tabular}[c]{@{}r@{}}0.736\\$\pm$0.011\end{tabular}}	& \multicolumn{1}{r|}{{\color[HTML]{0F2080} \textbf{\begin{tabular}[c]{@{}r@{}}0.923\\$\pm$0.004\end{tabular}}}}	& \multicolumn{1}{r|}{\begin{tabular}[c]{@{}r@{}}0.864\\$\pm$0.003\end{tabular}}									& \multicolumn{1}{r|}{\begin{tabular}[c]{@{}r@{}}0.964\\$\pm$0.005\end{tabular}}	& \multicolumn{1}{r|}{\begin{tabular}[c]{@{}r@{}}0.462\\$\pm$0.006\end{tabular}}									\\ \hline
		\textbf{Dropout} 															& \multicolumn{1}{l|}{{\begin{tabular}[c]{@{}l@{}}Architecture\end{tabular}}}    & \multicolumn{1}{r|}{{\color[HTML]{DF6747} \textbf{\begin{tabular}[c]{@{}r@{}}0.099\\$\pm$0.028\end{tabular}}}}													& \multicolumn{1}{r|}{\begin{tabular}[c]{@{}r@{}}0.470\\$\pm$0.120\end{tabular}}									& \multicolumn{1}{r|}{{\color[HTML]{DF6747} \textbf{\begin{tabular}[c]{@{}r@{}}0.477\\$\pm$0.035\end{tabular}}}}	& \multicolumn{1}{r|}{{\color[HTML]{DF6747} \textbf{\begin{tabular}[c]{@{}r@{}}0.644\\$\pm$0.073\end{tabular}}}}	& \multicolumn{1}{r|}{\begin{tabular}[c]{@{}r@{}}0.516\\$\pm$0.052\end{tabular}}									& \multicolumn{1}{r|}{\begin{tabular}[c]{@{}r@{}}0.739\\$\pm$0.016\end{tabular}}	& \multicolumn{1}{r|}{{\color[HTML]{0F2080} \textbf{\begin{tabular}[c]{@{}r@{}}0.925\\$\pm$0.003\end{tabular}}}}	& \multicolumn{1}{r|}{{\color[HTML]{0F2080} \textbf{\begin{tabular}[c]{@{}r@{}}0.867\\$\pm$0.002\end{tabular}}}}	& \multicolumn{1}{r|}{\begin{tabular}[c]{@{}r@{}}0.967\\$\pm$0.002\end{tabular}}	& \multicolumn{1}{r|}{\begin{tabular}[c]{@{}r@{}}0.459\\$\pm$0.008\end{tabular}}									\\ \hline\hline
		\textbf{\begin{tabular}[c]{@{}l@{}}Batch\\normalisation\end{tabular}} 		& \multicolumn{1}{l|}{{\begin{tabular}[c]{@{}l@{}}Training\end{tabular}}}        & \multicolumn{1}{r|}{{\color[HTML]{0F2080} \textbf{\begin{tabular}[c]{@{}r@{}}0.553\\$\pm$0.028\end{tabular}}}}													& \multicolumn{1}{r|}{\begin{tabular}[c]{@{}r@{}}0.604\\$\pm$0.048\end{tabular}}									& \multicolumn{1}{r|}{{\color[HTML]{0F2080} \textbf{\begin{tabular}[c]{@{}r@{}}0.784\\$\pm$0.006\end{tabular}}}}	& \multicolumn{1}{r|}{{\color[HTML]{0F2080} \textbf{\begin{tabular}[c]{@{}r@{}}0.911\\$\pm$0.004\end{tabular}}}}	& \multicolumn{1}{r|}{\begin{tabular}[c]{@{}r@{}}0.517\\$\pm$0.068\end{tabular}}									& \multicolumn{1}{r|}{\begin{tabular}[c]{@{}r@{}}0.739\\$\pm$0.010\end{tabular}}	& \multicolumn{1}{r|}{\begin{tabular}[c]{@{}r@{}}0.911\\$\pm$0.006\end{tabular}}									& \multicolumn{1}{r|}{\begin{tabular}[c]{@{}r@{}}0.865\\$\pm$0.005\end{tabular}}									& \multicolumn{1}{r|}{\begin{tabular}[c]{@{}r@{}}0.967\\$\pm$0.008\end{tabular}}	& \multicolumn{1}{r|}{\begin{tabular}[c]{@{}r@{}}0.469\\$\pm$0.009\end{tabular}}									\\ \hline
		\textbf{\begin{tabular}[c]{@{}l@{}}Layer\\normalisation\end{tabular}} 		& \multicolumn{1}{l|}{{\begin{tabular}[c]{@{}l@{}}Training\end{tabular}}}        & \multicolumn{1}{r|}{{\color[HTML]{0F2080} \textbf{\begin{tabular}[c]{@{}r@{}}0.531\\$\pm$0.025\end{tabular}}}}													& \multicolumn{1}{r|}{{\color[HTML]{DF6747} \textbf{\begin{tabular}[c]{@{}r@{}}0.217\\ $\pm$0.120\end{tabular}}}}	& \multicolumn{1}{r|}{{\color[HTML]{0F2080} \textbf{\begin{tabular}[c]{@{}r@{}}0.772\\$\pm$0.003\end{tabular}}}}	& \multicolumn{1}{r|}{{\color[HTML]{0F2080} \textbf{\begin{tabular}[c]{@{}r@{}}0.904\\$\pm$0.003\end{tabular}}}}  	& \multicolumn{1}{r|}{\begin{tabular}[c]{@{}r@{}}0.520\\$\pm$0.033\end{tabular}}									& \multicolumn{1}{r|}{\begin{tabular}[c]{@{}r@{}}0.722\\$\pm$0.021\end{tabular}}	& \multicolumn{1}{r|}{{\color[HTML]{DF6747} \textbf{\begin{tabular}[c]{@{}r@{}}0.903\\$\pm$0.005\end{tabular}}}}	& \multicolumn{1}{r|}{{\color[HTML]{0F2080} \textbf{\begin{tabular}[c]{@{}r@{}}0.869\\ $\pm$0.004\end{tabular}}}}	& \multicolumn{1}{r|}{\begin{tabular}[c]{@{}r@{}}0.965\\$\pm$0.006\end{tabular}}	& \multicolumn{1}{r|}{{\color[HTML]{0F2080} \textbf{\begin{tabular}[c]{@{}r@{}}0.581\\$\pm$0.012\end{tabular}}}}	\\ \hline
		\textbf{\begin{tabular}[c]{@{}l@{}}Weight\\normalisation\end{tabular}} 		& \multicolumn{1}{l|}{{\begin{tabular}[c]{@{}l@{}}Training\end{tabular}}}        & \multicolumn{1}{r|}{\begin{tabular}[c]{@{}r@{}}0.229\\$\pm$0.030\end{tabular}}																					& \multicolumn{1}{r|}{\begin{tabular}[c]{@{}r@{}}0.548\\$\pm$0.068\end{tabular}}									& \multicolumn{1}{r|}{\begin{tabular}[c]{@{}r@{}}0.731\\$\pm$0.037\end{tabular}}									& \multicolumn{1}{r|}{\begin{tabular}[c]{@{}r@{}}0.810\\$\pm$0.079\end{tabular}}									& \multicolumn{1}{r|}{\begin{tabular}[c]{@{}r@{}}0.515\\$\pm$0.042\end{tabular}}									& \multicolumn{1}{r|}{\begin{tabular}[c]{@{}r@{}}0.742\\$\pm$0.015\end{tabular}}	& \multicolumn{1}{r|}{\begin{tabular}[c]{@{}r@{}}0.911\\$\pm$0.004\end{tabular}}									& \multicolumn{1}{r|}{\begin{tabular}[c]{@{}r@{}}0.864\\$\pm$0.003\end{tabular}}									& \multicolumn{1}{r|}{\begin{tabular}[c]{@{}r@{}}0.965\\$\pm$0.006\end{tabular}}	& \multicolumn{1}{r|}{\begin{tabular}[c]{@{}r@{}}0.461\\$\pm$0.008\end{tabular}}									\\ \hline\hline
		\textbf{\begin{tabular}[c]{@{}l@{}}Regularisation\\term\end{tabular}} 		& \multicolumn{1}{l|}{{\begin{tabular}[c]{@{}l@{}}Loss function\end{tabular}}}	& \multicolumn{1}{r|}{{\color[HTML]{DF6747} \textbf{\begin{tabular}[c]{@{}r@{}}0.031\\$\pm$0.034\end{tabular}}}}													& \multicolumn{1}{r|}{\begin{tabular}[c]{@{}r@{}}0.549\\$\pm$0.066\end{tabular}}									& \multicolumn{1}{r|}{\begin{tabular}[c]{@{}r@{}}0.738\\$\pm$0.039\end{tabular}}									& \multicolumn{1}{r|}{{\color[HTML]{DF6747} \textbf{\begin{tabular}[c]{@{}r@{}}0.743\\$\pm$0.054\end{tabular}}}} 	& \multicolumn{1}{r|}{\begin{tabular}[c]{@{}r@{}}0.464\\$\pm$0.071\end{tabular}}									& \multicolumn{1}{r|}{\begin{tabular}[c]{@{}r@{}}0.731\\$\pm$0.025\end{tabular}}	& \multicolumn{1}{r|}{{\color[HTML]{0F2080} \textbf{\begin{tabular}[c]{@{}r@{}}0.918\\$\pm$0.004\end{tabular}}}}	& \multicolumn{1}{r|}{\begin{tabular}[c]{@{}r@{}}0.866\\$\pm$0.002\end{tabular}}									& \multicolumn{1}{r|}{\begin{tabular}[c]{@{}r@{}}0.963\\$\pm$0.010\end{tabular}}	& \multicolumn{1}{r|}{\begin{tabular}[c]{@{}r@{}}0.462\\$\pm$0.006\end{tabular}}									\\ \hline
    \end{tabular}
\end{table*}

The results obtained indicate that the performance of regularisation techniques differs across the datasets, and that not all techniques improve the generalisation performance over the baseline. Layer normalisation improved the performance over the baseline for five out of the ten datasets, but decreased the performance for two datasets. Batch normalisation increased the performance for three datasets, and had no significant effect on the other seven. Other techniques that provided limited improvement included pruning and regularisation term (improved for one dataset), and dropout (improved for two datasets).  In most cases, we see that the techniques either had no significant impact on results (such as weight normalisation), or even resulted in worse than the baseline performance for a number of datasets (such as weight perturbation). We also see that for some datasets, none of the implemented regularisation techniques resulted in a significant improvement in performance over the baseline. In the case of the Shoes dataset, for example, it is clear from Fig.~\ref{fig:ImagesCharts}(e) that techniques either had no impact, or resulted in worse performance than the baseline. 

The above leads to the conclusion that the impact of regularisation techniques on NN training is dependent on the dataset. The results we obtain for the batch normalisation are in line with Moradi et al.'s~\cite{moradi2020survey} study; however, our results for the geometric transformation, regularisation term, dropout and weight perturbation differed from their study (which was, however, limited to the Cifar10 dataset {and a specific NN architecture}).

It is also clear from the results that the effectiveness of regularisation techniques is dependent on the problem domain, due to the fact that some techniques only showed changes to the $F1$ scores of NNs trained on either the image problem domain or the numeric problem domain. Some examples of this are batch normalisation {(effective only on images)} and pruning {(effective only for numeric datasets)}. Other techniques, such as dropout and regularisation term, show a decrease in $F1$ scores of NNs trained in one problem domain, and an increase in the $F1$ scores of NNs trained in another problem domain. Still, other techniques, such as layer normalisation, cannot be grouped by problem domains.

{If we consider the four broad categories of regularisation, it appears from Table~\ref{tab:AvgAndStdOfF1Scores} that data-based strategies proved to be the least effective, and training-based strategies exhibited the highest rate of success. Training-based regularisation methods considered in the experiments were all based on the principle of normalisation, which may stabilise the training trajectory irrespective of the dataset. Data-based regularisation techniques, although known to be effective in the literature, may be especially dataset dependent, since each dataset exhibits specific distribution skews that may be corrected by one data-based technique but not another.}

\begin{figure*}[htbp]
    \centering
    \begin{subfigure}{0.3\textwidth}
        \centering
        \includegraphics[width=\linewidth]{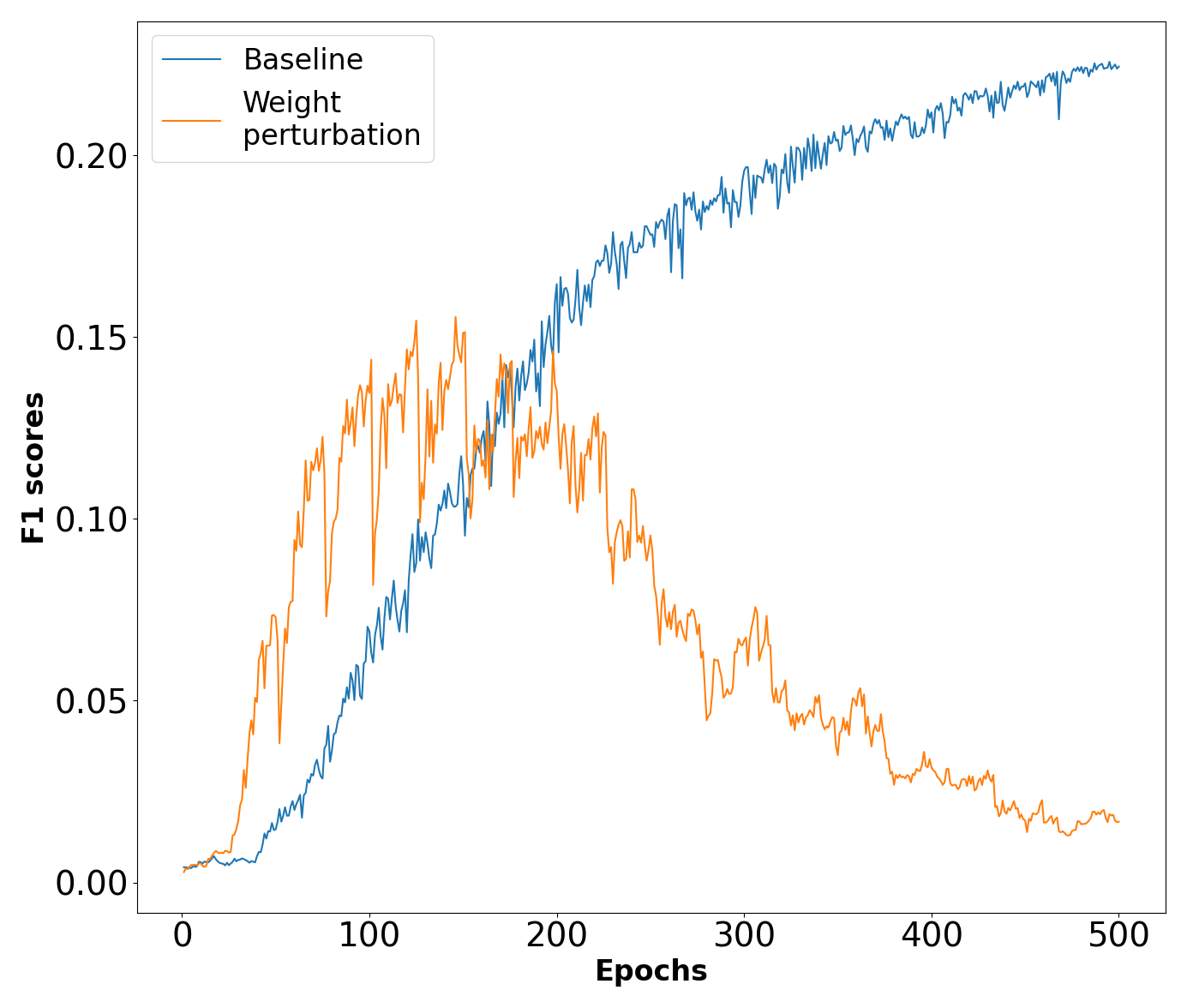}
        \caption{Balls Dataset}
    \end{subfigure}  
    \begin{subfigure}{0.3\textwidth}
        \centering
        \includegraphics[width=\linewidth]{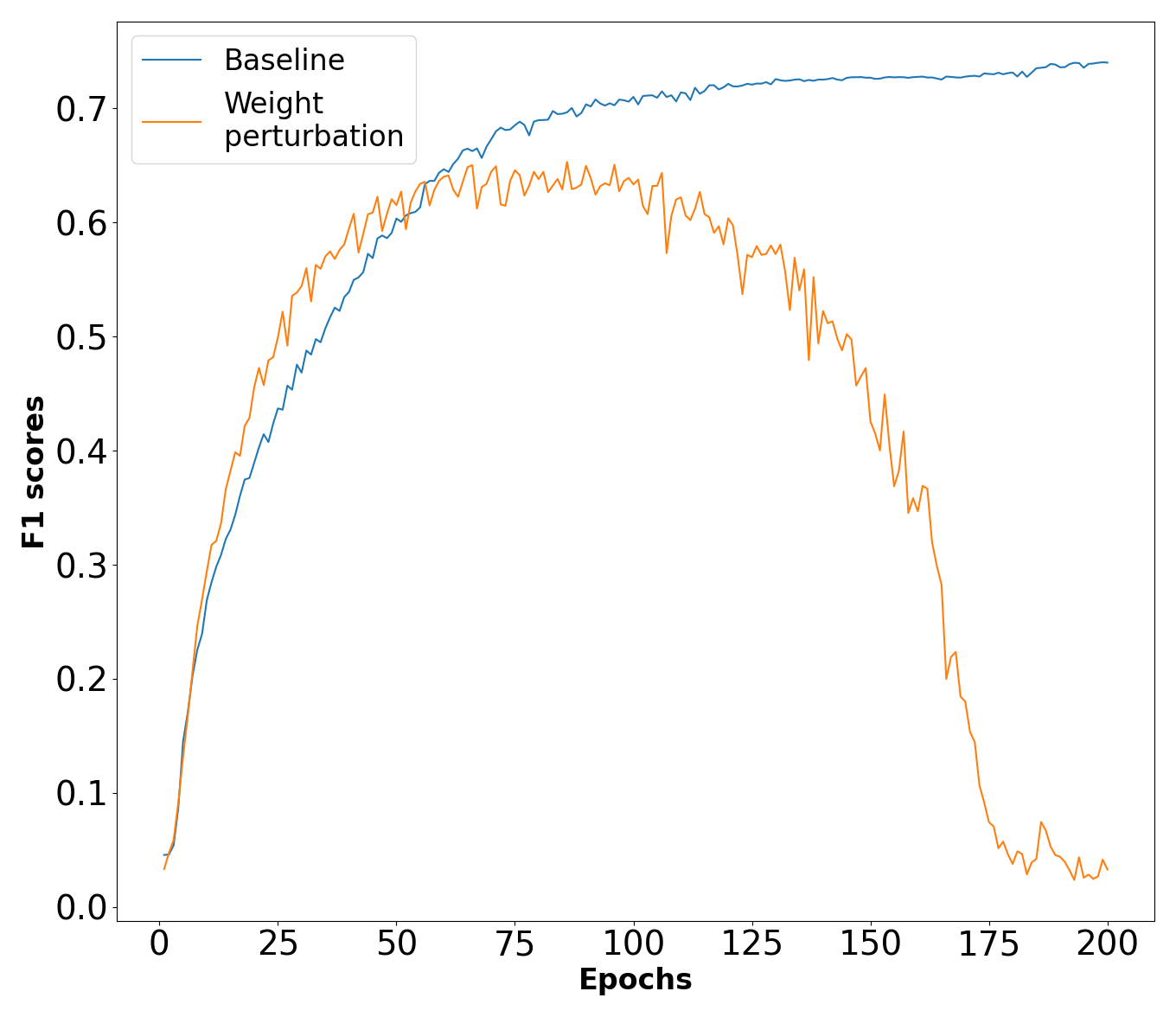}
        \caption{Cifar10 Dataset}
    \end{subfigure}  
    \begin{subfigure}{0.3\textwidth}
        \centering
        \includegraphics[width=\linewidth]{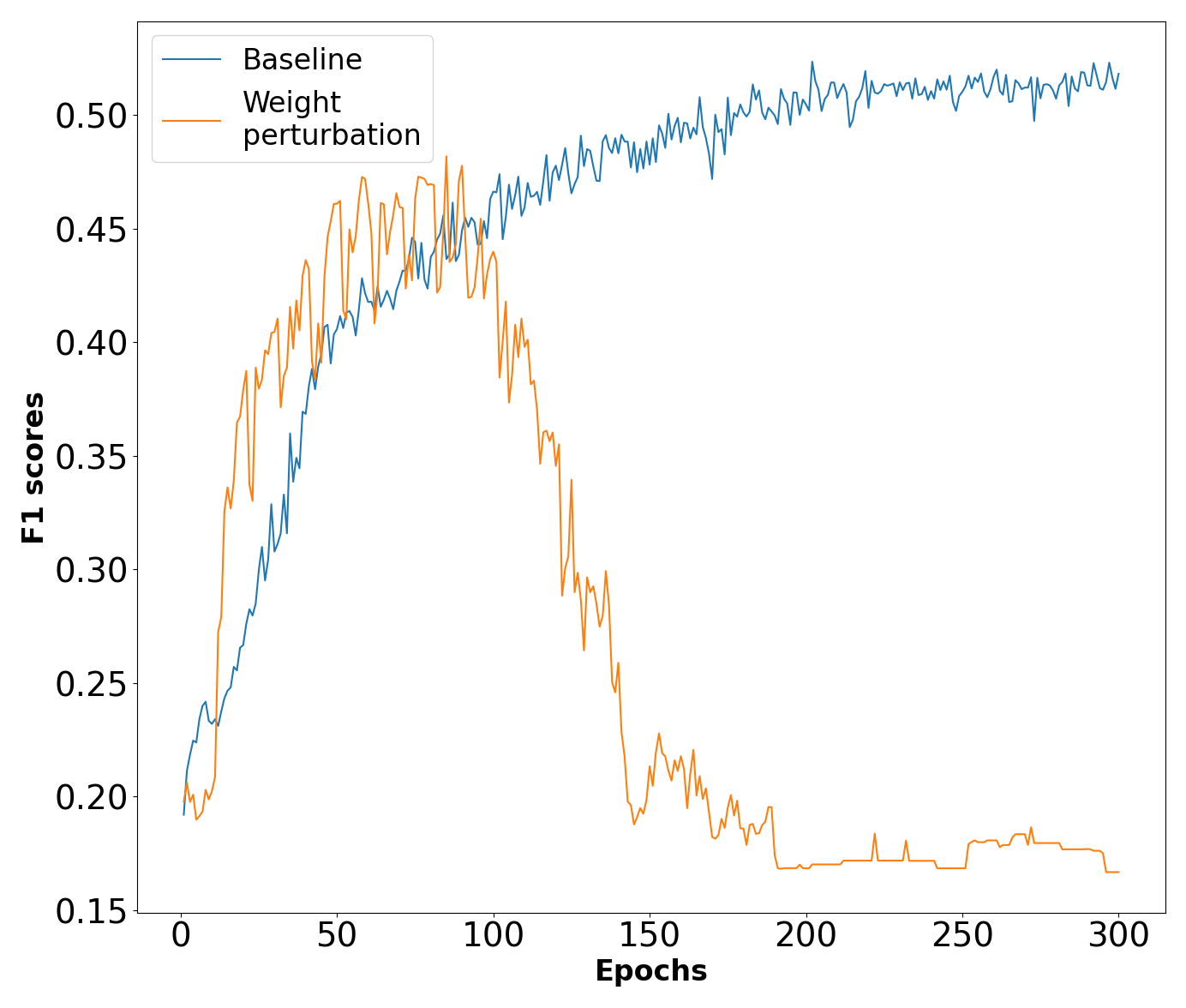}
        \caption{Shoes Dataset}
    \end{subfigure} 
    \caption{Line charts of the testing F1 scores for various image datasets vs epochs.}
    \label{fig:ImagesLineCharts}
\end{figure*}

Fig.~\ref{fig:ImagesLineCharts} shows that NN trained on the Balls, Cifar10 and Shoes datasets using weight perturbation reached higher $F1$ scores earlier than the baseline NN, after which the $F1$ score dropped off. Therefore, it can be concluded that weight perturbation leads to earlier improvement in generalisation, but is very destructive when implemented in later epochs during training. While the focus of this study is not on a specific regularisation technique, this observation confirms that applying regularisation methods blindly may in fact hinder rather than help generalisation performance. There is a clear gap in our understanding of how the dataset and the NN architecture affect regularisation effectiveness, and what phases of training would benefit from the various regularisation methods the most. We hope that future research can shed light on these important questions.

\section{{Summary}}\label{sec:conclusion}
This paper proposed a taxonomy of regularisation techniques under four broad categories, namely: data-based {strategies} (techniques that manipulate the training data or the latent space representation of the training data), architecture {strategies} (techniques that manipulate the architecture of a NN), training {strategies} (techniques that manipulate the training process of a NN), and loss function {strategies} (techniques that manipulate the loss function of a NN). This taxonomy provides a holistic view of regularisation techniques crucial to machine learning. 

We identified several contradictions and correspondences between families of regularisation techniques. The following families of techniques were found to contradict each other: early stopping and over-training, double descent and data augmentation or noise injection, and pruning and over-parameterisation. Meanwhile, the following families of techniques correspond with each other: dataset noise injection and data augmentation, dataset noise injection and regularisation penalty term, dropout and pruning, transfer learning and pre-training, and pre-training and pruning. Future research can investigate these contradictions and correspondences, and the impact that different families of techniques have on the loss landscape of a NN. 

In addition, we performed benchmark tests on the following nine regularisation techniques: batch normalisation, dropout, geometric transformation, layer normalisation, pruning, regularisation terms, SMOTE, weight normalisation, and weight perturbation. Benchmarking was conducted on five numeric datasets and five image datasets, and revealed that there is no one regularisation technique that works for all datasets. {Weight perturbation was the only technique that never helped and sometimes hindered; batch normalisation was the only technique than never hindered and sometimes helped. Further, we noted that training-based methods yielded a higher success rate than data-based methods in terms of positive effect on regularisation.} Future research will investigate the link between dataset characteristics and generalisation technique performance with the aim of identifying the most appropriate technique to use based on the meta-features of the dataset.



\bibliographystyle{IEEEtran}
\bibliography{references}

@article{zhu2025comprehensive,
  title={A comprehensive review of network pruning based on pruning granularity and pruning time perspectives},
  author={Zhu, Kehan and Hu, Fuyi and Ding, Yuanbing and Zhou, Wei and Wang, Ruxin},
  journal={Neurocomputing},
  pages={129382},
  year={2025},
  volume = {626},
  publisher={Elsevier}
}

@article(altarabichi2024rolling,
    title = "Rolling the dice for better deep learning performance: A study of randomness techniques in deep neural networks",
    journal = "Information Sciences",
    volume = "667",
    pages = "120500",
    year = "2024",
    issn = "0020-0255",
    doi = "https://doi.org/10.1016/j.ins.2024.120500",
    author = "Altarabichi, Mohammed Ghaith and Nowaczyk, Sławomir and Pashami, Sepideh and {Sheikholharam Mashhadi}, Peyman and Handl, Julia"
)

@article(an1996effects,
    author = "An, Guozhong",
    title = "The Effects of Adding Noise During Backpropagation Training on a Generalization Performance",
    journal = "Neural Computation",
    volume = "8",
    number = "3",
    pages = "643-674",
    year = "1996",
    month = "04",
    issn = "0899-7667",
    doi = "10.1162/neco.1996.8.3.643",
    eprint = "https://direct.mit.edu/neco/article-pdf/8/3/643/813312/neco.1996.8.3.643.pdf",
)

@inproceedings(arslan2019smote,
  title="{SMOTE} and {Gaussian} {Noise} {Based} {Sensor} {Data} {Augmentation}",
  author="Arslan, Mehmet and Guzel, Metehan and Demirci, Mehmet and Ozdemir, Suat",
  booktitle="2019 4th International Conference on Computer Science and Engineering (UBMK)",
  pages="1--5",
  year=2019,
  organization="IEEE"
)

@article(bartoldson2020generalization,
  title="The {Generalization-stability} {Trade-off} in {Neural} {Network} {Pruning}",
  author="Bartoldson, Brian and Morcos, Ari and Barbu, Adrian and Erlebacher, Gordon",
  journal="{A}dvances in {N}eural {I}nformation {P}rocessing {S}ystems",
  volume=33,
  pages="20852--20864",
  year=2020
)

@misc(bao2024recent,
      title="A Recent Survey of Heterogeneous Transfer Learning", 
      author="Runxue, Bao and Yiming, Sun and Yuhe, Gao and Jindong, Wang and Qiang, Yang and Zhi-Hong, Mao and Ye, Ye",
      year=2024,
      eprint="2310.08459",
      archivePrefix="arXiv",
      primaryClass="cs.LG",
      url="https://arxiv.org/abs/2310.08459", 
)

@article(belkin2019reconciling,
  title="Reconciling {Modern} {Machine-learning} {Practice} and the {Classical} {Bias-variance} {Trade-off}",
  author="Belkin, Mikhail and Hsu, Daniel and Ma, Siyuan and Mandal, Soumik",
  journal="Proceedings of the National Academy of Sciences",
  volume=116,
  number=32,
  pages="15849--15854",
  year=2019,
  publisher="National Acad Sciences"
)

@article(belkin2020two,
  title="Two {Models} of {Double} {Descent} for {Weak} {Features}",
  author="Belkin, Mikhail and Hsu, Daniel and Xu, Ji",
  journal="SIAM Journal on Mathematics of Data Science",
  volume=2,
  number=4,
  pages="1167--1180",
  year=2020,
  publisher="SIAM"
)

@article(bishop1995training,
  title="Training with {Noise} is {Equivalent} to {Tikhonov} {Regularization}",
  author="Bishop, Chris M",
  journal="Neural Computation",
  volume=7,
  number=1,
  pages="108--116",
  year=1995,
  publisher="MIT Press"
)

@misc(bisla2022lowpass,
    title="{L}ow-{P}ass {F}iltering {SGD} for {R}ecovering {F}lat {O}ptima in the {D}eep {L}earning Optimization Landscape", 
    author="Devansh Bisla and Jing Wang and Anna Choromanska",
    year="2022",
    eprint="2201.08025",
    archivePrefix="arXiv",
    primaryClass="cs.LG",
    url="https://arxiv.org/abs/2201.08025",
)

@inProceedings(blalock2020state,
  title="What is the {State} of {Neural} {Network} {Pruning}?",
  author="Blalock, Davis and Gonzalez Ortiz, Jose Javier and Frankle, Jonathan and Guttag, John",
  booktitle="Proceedings of Machine Learning and Systems",
  volume=2,
  pages="129--146",
  year=2020
)

@MISC(Bock2007Magic,
  title    = "Magic {G}amma {T}elescope",
  author   = "Bock, R",
  year     =  2004,
  howpublished = "UCI Machine Learning Repository",
  language = "en",
  URL       = "https://doi.org/10.24432/C52C8B"
)

@article(bosman2020visualising,
  title="Visualising {Basins} of {Attraction} for the {Cross-entropy} and the {Squared} {Error} {Neural} {Network} {Loss} {Functions}",
  author="Bosman, Anna Sergeevna and Engelbrecht, Andries and Helbig, Mard{\'e}",
  journal="Neurocomputing",
  volume=400,
  pages="113--136",
  year=2020,
  publisher="Elsevier"
)

@article(caruana2000overfitting,
  title="Overfitting in {Neural} {Nets}: {Backpropagation}, {Conjugate} {Gradient}, and {Early} {Stopping}",
  author="Caruana, Rich and Lawrence, Steve and Giles, C",
  journal="{A}dvances in {N}eural {I}nformation {P}rocessing {S}ystems",
  volume=13,
  year=2000,
pages="381 - 387"
)

@article(chaudhari2019entropy,
  title="Entropy-sgd: Biasing gradient descent into wide valleys",
  author="Chaudhari, Pratik and Choromanska, Anna and Soatto, Stefano and LeCun, Yann and Baldassi, Carlo and Borgs, Christian and Chayes, Jennifer and Sagun, Levent and Zecchina, Riccardo",
  journal="Journal of Statistical Mechanics: Theory and Experiment",
  volume=2019,
  number=12,
  pages=124018,
  year=2019,
  publisher="IOP Publishing"
)

@inproceedings(chen2021lottery,
  title="The {Lottery} {Tickets} {Hypothesis} for {Supervised} and {Self-supervised} {Pre-training} in {Computer} {Vision} {Models}",
  author="Chen, Tianlong and Frankle, Jonathan and Chang, Shiyu and Liu, Sijia and Zhang, Yang and Carbin, Michael and Wang, Zhangyang",
  booktitle="Proceedings of the IEEE/CVF {C}onference on {C}omputer {V}ision and {P}attern {R}ecognition",
  pages="16306--16316",
  year=2021
)

@inproceedings(chen2020simple,
  title="A {S}imple {F}ramework for {C}ontrastive {L}earning of {V}isual {Representations}",
  author="Chen, Ting and Kornblith, Simon and Norouzi, Mohammad and Hinton, Geoffrey",
  booktitle="International {C}onference on {M}achine {L}earning",
  pages="1597--1607",
  year=2020,
  organization="PMLR"
)

@article(Cortez2009ModelingWP,
  title="Modeling wine preferences by data mining from physicochemical properties",
  author="P. Cortez and Antonio Lu{\'i}z Cerdeira and Fernando Almeida and Telmo Matos and Jos{\'e} Reis",
  journal="Decis. Support Syst.",
  year="2009",
  volume="47",
  pages="547-553"
)

@inproceedings(cubuk2018autoaugment,
  title="Autoaugment: Learning {A}ugmentation {S}trategies {F}rom {D}ata",
  author="Cubuk, Ekin D and Zoph, Barret and Mane, Dandelion and Vasudevan, Vijay and Le, Quoc V",
  booktitle="Proceedings of the IEEE/CVF {C}onference on {C}omputer {V}ision and {P}attern {R}ecognition",
  pages="113--123",
  year=2019
)

@inproceedings(devries2017dataset,
      title="{D}ataset {A}ugmentation in {F}eature {S}pace", 
      author="Terrance DeVries and Graham W. Taylor",
      year=2017,
      eprint="1702.05538",
      archivePrefix="arXiv",
      primaryClass="stat.ML",
      url="https://arxiv.org/abs/1702.05538" 
)

@inproceedings(dickson2021hybridised,
  title="Hybridised {L}oss {F}unctions for {I}mproved {N}eural {N}etwork {G}eneralisation",
  author="Dickson, Matthew C and Bosman, Anna S and Malan, Katherine M",
  booktitle="Pan-African Artificial Intelligence and Smart Systems Conference",
  pages="169--181",
  year=2021,
  organization="Springer"
)

@MISC(Dickson2023Cirrhosis,
  title    = "Cirrhosis {P}atient {S}urvival {P}rediction",
  author   = "Dickson, E and Grambsch, P and Fleming, T and Fisher, L and
              Langworthy, A",
  year     =  1989,
  howpublished = "UCI Machine Learning Repository",
  language = "en",
  URL = "https://doi.org/10.24432/C5R02G"
)

@article(dziugaite2017computing,
  title="Computing nonvacuous generalization bounds for deep (stochastic) neural networks with many more parameters than training data",
  author="Dziugaite, Gintare Karolina and Roy, Daniel M",
  journal="arXiv preprint arXiv:1703.11008",
  year=2017
)

@inproceedings(erhan2009difficulty,
  title="The {D}ifficulty of {T}raining {D}eep {A}rchitectures and the {E}ffect of {U}nsupervised {P}re-training",
  author="Erhan, Dumitru and Manzagol, Pierre-Antoine and Bengio, Yoshua and Bengio, Samy and Vincent, Pascal",
  booktitle="Artificial {I}ntelligence and {S}tatistics",
  pages="153--160",
  year=2009,
  organization="PMLR"
)

@article(ferro2023early,
  title="Early stopping by correlating online indicators in neural networks",
  author="Ferro, Manuel Vilares and Mosquera, Yerai Doval and Pena, Francisco J Ribadas and Bilbao, V{\'\i}ctor M Darriba",
  journal="Neural Networks",
  volume=159,
  pages="109--124",
  year=2023,
  publisher="Elsevier"
)

@inproceedings(frankle2018lottery,
  title="The {L}ottery {T}icket {H}ypothesis: Finding {S}parse, {T}rainable {N}eural {N}etworks",
  author="Frankle, Jonathan and Carbin, Michael",
  booktitle="ICLR",
  year=2019
)

@misc(foret2021sharpness,
    title="Sharpness-{A}ware {M}inimization for {E}fficiently {I}mproving {G}eneralization", 
    author="Pierre Foret and Ariel Kleiner and Hossein Mobahi and Behnam Neyshabur",
    year=2021,
    eprint="2010.01412",
    archivePrefix="arXiv",
    primaryClass="cs.LG",
    url="https://arxiv.org/abs/2010.01412"
)

@article(ganin2016domain,
  title="Domain-adversarial {T}raining of {N}eural {N}etworks",
  author="Ganin, Yaroslav and Ustinova, Evgeniya and Ajakan, Hana and Germain, Pascal and Larochelle, Hugo and Laviolette, Fran{\c{c}}ois and Marchand, Mario and Lempitsky, Victor",
  journal="Journal of Machine Learning Research",
  volume=17,
  number=1,
  pages="2096--2030",
  year=2016,
  publisher="JMLR. org"
)

@inproceedings(gao2022loss,
  title="Loss {F}unction {L}earning for {D}omain {G}eneralization by {I}mplicit {G}radient",
  author="Gao, Boyan and Gouk, Henry and Yang, Yongxin and Hospedales, Timothy",
  booktitle="International Conference on Machine Learning",
  pages="7002--7016",
  year=2022,
  organization="PMLR"
)

@article(ghorbanali2022ensemble,
  title="Ensemble {T}ransfer {L}earning-based {M}ultimodal {S}entiment {A}nalysis {U}sing {W}eighted {C}onvolutional {N}eural {N}etworks",
  author="Ghorbanali, Alireza and Sohrabi, Mohammad Karim and Yaghmaee, Farzin",
  journal="Information Processing \& Management",
  volume=59,
  number=3,
  pages=102929,
  year=2022,
  publisher="Elsevier"
)

@inproceedings(gitman2017comparison,
  title="Comparison of {B}atch {N}ormalization and {W}eight {N}ormalization {A}lgorithms for the {L}arge-scale {I}mage {C}lassification",
      author="Igor Gitman and Boris Ginsburg",
      year=2017,
      eprint="1709.08145",
      archivePrefix="arXiv",
      primaryClass="cs.CV",
      url="https://arxiv.org/abs/1709.08145"
)

@inproceedings(golik2013cross,
  title="Cross-entropy vs. {S}quared {E}rror {T}raining: {A} {T}heoretical and {E}xperimental {C}omparison.",
  author="Golik, Pavel and Doetsch, Patrick and Ney, Hermann",
  booktitle="Interspeech",
  volume=13,
  pages="1756--1760",
  year=2013
)

@inproceedings(gonzalez2020improved,
  title="Improved {T}raining {S}peed, {A}ccuracy, and {D}ata {U}tilization {T}hrough {L}oss {F}unction {O}ptimization",
  author="Gonzalez, Santiago and Miikkulainen, Risto",
  booktitle="2020 IEEE Congress on Evolutionary Computation (CEC)",
  pages="1--8",
  year=2020,
)

@book(goodfellow2016deep,
  title="Deep {L}earning",
  author="Goodfellow, Ian and Bengio, Yoshua and Courville, Aaron",
  year=2016,
  publisher="MIT {P}ress"
)

@misc(gomez2019learning,
      title="Learning {S}parse {N}etworks {U}sing {T}argeted {D}ropout",
      author="Gomez, Aidan N and Zhang, Ivan and Kamalakara, Siddhartha Rao and Madaan, Divyam and Swersky, Kevin and Gal, Yarin and Hinton, Geoffrey E",
      year=2019,
      eprint="1905.13678",
      archivePrefix="arXiv",
      primaryClass="cs.LG",
      url="https://arxiv.org/abs/1905.13678", 
)

@article(Han2017Fashion,
  author       = "Han Xiao and Kashif Rasul and Roland Vollgraf",
  title        = "Fashion-MNIST: a Novel Image Dataset for Benchmarking Machine Learning Algorithms",
  journal      = "CoRR",
  volume       = "abs/1708.07747",
  year         = "2017",
  eprinttype    = "arXiv",
  eprint       = "1708.07747",
  bibsource    = "dblp computer science bibliography, https://dblp.org"
)

@article(han2015learning,
  title="Learning {B}oth {W}eights and {C}onnections for {E}fficient {N}eural {N}etwork",
  author="Han, Song and Pool, Jeff and Tran, John and Dally, William",
  journal="{A}dvances in {N}eural {I}nformation {P}rocessing {S}ystems",
  volume=28,
  year=2015,
  pages="1135 - 1143"
)

@article(han2018batch,
  title="Batch-normalized {MLPconv-wise} {S}upervised {P}re-training {N}etwork in {N}etwork",
  author="Han, Xiaomeng and Dai, Qun",
  journal="Applied Intelligence",
  volume=48,
  pages="142--155",
  year=2018,
  publisher="Springer"
)

@book(hastie2009elements,
  title="The {E}lements of {S}tatistical {L}earning: {D}ata {M}ining, {I}nference, and {P}rediction",
  author="Hastie, Trevor and Tibshirani, Robert and Friedman, Jerome H",
  year=2009,
  publisher="Springer",
  address="New York, NY",
  edition="2nd",
  isbn="978-0-387-84856-5"
)

@inproceedings(hinton2012improving,
    title="Improving {N}eural {N}etworks by {P}reventing {C}o-adaptation of {F}eature {D}etectors", 
    author="Geoffrey E. Hinton and Nitish Srivastava and Alex Krizhevsky and Ilya Sutskever and Ruslan R. Salakhutdinov",
    year=2012,
    eprint="1207.0580",
    archivePrefix="arXiv",
    primaryClass="cs.NE",
    url="https://arxiv.org/abs/1207.0580"
)

@article(huang2006correcting,
  title="Correcting {S}ample {S}election {B}ias by {U}nlabeled {D}ata",
  author="Huang, Jiayuan and Gretton, Arthur and Borgwardt, Karsten and Sch{\"o}lkopf, Bernhard and Smola, Alex",
  abstarct="We consider the scenario where training and test data are drawn from different distributions, commonly referred to as sample selection bias. Most algorithms for this setting try to ﬁrst recover sampling distributions and then make appropriate corrections based on the distribution estimate. We present a nonparametric method which directly produces resampling weights without distribution estimation. Our method works by matching distributions between training and testing sets in feature space. Experimental results demonstrate that our method works well in practice.",
  journal="{A}dvances in {N}eural {I}nformation {P}rocessing {S}ystems",
  volume=19,
  year=2006,
  pages="601 - 608"
)

@inproceedings(huang2018orthogonal,
  title="Orthogonal {W}eight {N}ormalization: {S}olution to {O}ptimization {O}ver {M}ultiple {D}ependent {S}tiefel {M}anifolds in {D}eep {N}eural {N}etworks",
  author="Huang, Lei and Liu, Xianglong and Lang, Bo and Yu, Adams and Wang, Yongliang and Li, Bo",
  abstarct="Orthogonal matrix has shown advantages in training Recurrent Neural Networks (RNNs), but such matrix is limited to be square for the hidden-to-hidden transformation in RNNs. In this paper, we generalize such square orthogonal matrix to orthogonal rectangular matrix and formulating this problem in feed-forward Neural Networks (FNNs) as Optimization over Multiple Dependent Stiefel Manifolds (OMDSM). We show that the orthogonal rectangular matrix can stabilize the distribution of network activations and regularize FNNs. We propose a novel orthogonal weight normalization method to solve OMDSM. Particularly, it constructs orthogonal transformation over proxy parameters to ensure the weight matrix is orthogonal. To guarantee stability, we minimize the distortions between proxy parameters and canonical weights over all tractable orthogonal transformations. In addition, we design orthogonal linear module (OLM) to learn orthogonal filter banks in practice, which can be used as an alternative to standard linear module. Extensive experiments demonstrate that by simply substituting OLM for standard linear module without revising any experimental protocols, our method improves the performance of the state-of-the-art networks, including Inception and residual networks on CIFAR and ImageNet datasets.",
  booktitle="Proceedings of the AAAI Conference on Artificial Intelligence",
  volume=32,
  number=1,
  year=2018,
    pages="3271 - 3278"
)

@inproceedings(huang2017centered,
  title="Centered {W}eight {N}ormalization in {A}ccelerating {T}raining of {D}eep {N}eural {N}etworks",
  author="Huang, Lei and Liu, Xianglong and Liu, Yang and Lang, Bo and Tao, Dacheng",
  abstarct="Training deep neural networks is difficult for the pathological curvature problem. Re-parameterization is an effective way to relieve the problem by learning the curvature approximately or constraining the solutions of weights with good properties for optimization. This paper proposes to re-parameterize the input weight of each neuron in deep neural networks by normalizing it with zero-mean and unit-norm, followed by a learnable scalar parameter to adjust the norm of the weight. This technique effectively stabilizes the distribution implicitly. Besides, it improves the conditioning of the optimization problem and thus accelerates the training of deep neural networks. It can be wrapped as a linear module in practice and plugged in any architecture to replace the standard linear module. We highlight the benefits of our method on both multi-layer perceptrons and convolutional neural networks, and demonstrate its scalability and efficiency on SVHN, CIFAR-10, CIFAR-100 and ImageNet datasets.
",
  booktitle="Proceedings of the IEEE International Conference on Computer Vision",
  pages="2803--2811",
  year=2017
)

@article(huisman2021survey,
  title="A {S}urvey of {D}eep {M}eta-learning",
  author="Huisman, Mike and Van Rijn, Jan N and Plaat, Aske",
  abstarct="Deep neural networks can achieve great successes when presented with large data sets and sufficient computational resources. However, their ability to learn new concepts quickly is limited. Meta-learning is one approach to address this issue, by enabling the network to learn how to learn. The field of Deep Meta-Learning advances at great speed, but lacks a unified, in-depth overview of current techniques. With this work, we aim to bridge this gap. After providing the reader with a theoretical foundation, we investigate and summarize key methods, which are categorized into (i) metric-, (ii) model-, and (iii) optimization-based techniques. In addition, we identify the main open challenges, such as performance evaluations on heterogeneous benchmarks, and reduction of the computational costs of meta-learning.",
  journal="Artificial Intelligence Review",
  volume=54,
  number=6,
  pages="4483--4541",
  year=2021,
  publisher="Springer"
)

@article(hochreiter1994simplifying,
  title="Simplifying neural nets by discovering flat minima",
  author="Hochreiter, Sepp and Schmidhuber, J{\"u}rgen",
  abstarct="We present a new algorithm for finding low complexity networks with high generalization capability. The algorithm searches for large connected regions of so-called''fiat''minima of the error func (cid: 173) tion. In the weight-space environment of a 'flat' minimum, the error remains approximately constant. Using an MDL-based ar (cid: 173) gument, flat minima can be shown to correspond to low expected overfitting. Although our algorithm requires the computation of second order derivatives, it has backprop's order of complexity. Experiments with feedforward and recurrent nets are described. In an application to stock market prediction, the method outperforms conventional backprop, weight decay, and 'optimal brain surgeon'.",
  journal="{A}dvances in {N}eural {I}nformation {P}rocessing {S}ystems",
  volume=7,
  year=1994,
pages="529 - 536"
)

@article(hochreiter1997flat,
  title="Flat minima",
  author="Hochreiter, Sepp and Schmidhuber, J{\"u}rgen",
  abstarct="We present a new algorithm for finding low-complexity neural networks with high generalization capability. The algorithm searches for a “flat” minimum of the error function. A flat minimum is a large connected region in weight space where the error remains approximately constant. An MDL-based, Bayesian argument suggests that flat minima correspond to “simple” networks and low expected overfitting. The argument is based on a Gibbs algorithm variant and a novel way of splitting generalization error into underfitting and overfitting error. Unlike many previous approaches, ours does not require Gaussian assumptions and does not depend on a “good” weight prior. Instead we have a prior over input output functions, thus taking into account net architecture and training set. Although our algorithm requires the computation of second-order derivatives, it has backpropagation's order of complexity. Automatically, it effectively prunes units, weights, and input lines. Various experiments with feedforward and recurrent nets are described. In an application to stock market prediction, flat minimum search outperforms conventional backprop, weight decay, and “optimal brain surgeon/optimal brain damage.”",
  journal="Neural Computation",
  volume=9,
  number=1,
  pages="1--42",
  year=1997,
  publisher="MIT Press One Rogers Street, Cambridge, MA 02142-1209, USA journals-info~…"
)

@article(holmstrom1992using,
  title="Using {A}dditive {N}oise in {B}ack-propagation {T}raining",
  author="Holmstrom, Lasse and Koistinen, Petri",
  journal="IEEE Transactions on Neural Networks",
  volume=3,
  number=1,
  pages="24--38",
  year=1992
)

@ARTICLE(huang2023normalization,
  title="Normalization {T}echniques in {T}raining {DNNs}: {M}ethodology, {A}nalysis and {A}pplication",
  journal ="{IEEE} {T}ransactions on {P}attern {A}nalysis and {M}achine {I}ntelligence",  
  year=2023,
  volume=45,
  number=8,
  pages="10173-10196",
  doi="10.1109/TPAMI.2023.3250241"
)

@MISC(iron2022nike,
  title    = "{N}ike, {A}didas and {C}onverse {S}hoes {I}mages",
  author   = "Iron486",
  year     =  2022,
  language = "en",
  howpublished = "Kaggle Repository",
  URL       = "https://www.kaggle.com/datasets/die9origephit/nike-adidas-and-converse-imaged/data", 
    note        = "Accessed: 03 January 2024"
)

@inproceedings(inayoshi2005improved,
  title="Improved {G}eneralization by {A}dding {B}oth {A}uto-association and {H}idden-layer-noise to {N}eural-network-based-classifiers",
  author="Inayoshi, Hiroaki and Kurita, Takio",
  booktitle="2005 IEEE Workshop on Machine Learning for Signal Processing",
  pages="141--146",
  year=2005,
  organization="IEEE"
)

@inproceedings(ioffe2015batch,
  title="Batch {N}ormalization: Accelerating {D}eep {N}etwork {T}raining by {R}educing {I}nternal {C}ovariate {S}hift",
  author="Ioffe, Sergey and Szegedy, Christian",
  booktitle="International {C}onference on {M}achine {L}earning",
  pages="448--456",
  year=2015,
  organization="PMLR"
)

@inproceedings(iyer2000novel,
  title="A {N}ovel {M}ethod to {S}top {N}eural {N}etwork {T}raining",
  author="Iyer, Mahesh S and Rhinehart, R Russell",
  booktitle="Proceedings of the 2000 American Control Conference. ACC (IEEE Cat. No. 00CH36334)",
  volume=2,
  pages="929--933",
  year=2000,
  organization="IEEE"
)

@misc(izmailov2019averaging,
    title="Averaging {W}eights {L}eads to {W}ider {O}ptima and {B}etter {G}eneralization", 
    author="Pavel Izmailov and Dmitrii Podoprikhin and Timur Garipov and Dmitry Vetrov and Andrew Gordon Wilson",
    year=2019,
    eprint="1803.05407",
    archivePrefix="arXiv",
    primaryClass="cs.LG",
    url="https://arxiv.org/abs/1803.05407"
)

@article(jin2022pruning,
  title="Pruning’s {E}ffect on {G}eneralization {T}hrough the {L}ens of {T}raining and {R}egularization",
  author="Jin, Tian and Carbin, Michael and Roy, Dan and Frankle, Jonathan and Dziugaite, Gintare Karolina",
  journal="{A}dvances in {N}eural {I}nformation {P}rocessing {S}ystems",
  volume=35,
  pages="37947--37961",
  year=2022
)

@misc(jin2024survey,
      title="A {S}urvey on {M}ixup {A}ugmentations and {B}eyond", 
      author="Xin Jin and Hongyu Zhu and Siyuan Li and Zedong Wang and Zicheng Liu and Chang Yu and Huafeng Qin and Stan Z. Li",
      year=2024,
      eprint="2409.05202",
      archivePrefix="arXiv",
      primaryClass="cs.LG",
      url="https://arxiv.org/abs/2409.05202"
)

@inproceedings(kaddour2022when,
    title="When Do Flat Minima Optimizers Work?",
    author="Jean Kaddour and Linqing Liu and Ricardo Silva and Matt Kusner",
    booktitle="Advances in Neural Information Processing Systems",
    editor="Alice H. Oh and Alekh Agarwal and Danielle Belgrave and Kyunghyun Cho",
    year="2022",
    pages="16577-16595"
)

@MISC(Kahn2023Diabetes,
  author       = "Kahn, Michael",
  title        = "Diabetes",
  howpublished = "UCI Machine Learning Repository",
  url         = "https://doi.org/10.24432/C5T59G"
)

@inbook(kawaguchi2022generalization, 
    place="Cambridge", 
    title="Generalization in Deep Learning", 
    booktitle="Mathematical Aspects of Deep Learning", 
    publisher="Cambridge University Press", 
    author="Kawaguchi, K. and Bengio, Y. and Kaelbling, L.", 
    editor="Grohs, Philipp and Kutyniok, GittaEditors", 
    year=2022, 
    pages="112--148"
)

@article(kukavcka2017regularization,
  title="Regularization for {D}eep {L}earning: A {T}axonomy",
  author="Kuka{\v{c}}ka, Jan and Golkov, Vladimir and Cremers, Daniel",
  journal="arXiv preprint arXiv:1710.10686",
  year=2017,
  volume="abs/1710.10686",
  url="https://api.semanticscholar.org/CorpusID:8689292"
)

@TECHREPORT(Krizhevsky2009Learning,
  title    = "Learning Multiple Layers of Features from Tiny Images",
  author   = "Krizhevsky, Alex",
  month    =  apr,
  year     =  2009,
  language = "en",
  URL      = "https://www.cs.toronto.edu/~kriz/learning-features-2009-TR.pdf",
  note        = "Accessed: 03 January 2024"
)

@article(lechner2022pyhopper,
	title="PyHopper -- Hyperparameter optimization", 
	author="Mathias, Lechner and Ramin, Hasani and Philipp, Neubauer and Sophie, Neubauer and Daniela, Rus",
	journal="arXiv preprint arXiv:2210.04728",
	year=2022
)

@article(lee2018snip,
  title="{SNIP}: Single-shot network pruning based on connection sensitivity",
  author="Lee, Namhoon and Ajanthan, Thalaiyasingam and Torr, Philip HS",
  journal="arXiv preprint arXiv:1810.02340",
  year=2018
)

@article(lemley2017smart,
  title="Smart {A}ugmentation {L}earning an {O}ptimal {D}ata {A}ugmentation {S}trategy",
  author="Lemley, Joseph and Bazrafkan, Shabab and Corcoran, Peter",
  journal="IEEE Access",
  volume=5,
  pages="5858--5869",
  year=2017,
  publisher="IEEE"
)

@misc(li2016pruning,
  title="Pruning filters for efficient convnets",
  author="Li, Hao and Kadav, Asim and Durdanovic, Igor and Samet, Hanan and Graf, Hans Peter",
  year=2017,
  eprint="1608.08710",
  archivePrefix="arXiv",
  primaryClass="cs.CV",
  url="https://arxiv.org/abs/1608.08710" 
)

@article(liu2008optimized,
  title="Optimized {A}pproximation {A}lgorithm in {N}eural {N}etworks {W}ithout {O}verfitting",
  author="Liu, Yinyin and Starzyk, Janusz A and Zhu, Zhen",
  journal="IEEE Transactions on Neural Networks",
  volume=19,
  number=6,
  pages="983--995",
  year=2008,
  publisher="IEEE"
)

@inproceedings(lodwich2009evaluation,
  title="Evaluation of {R}obustness and {P}erformance of {E}arly {S}topping {R}ules with {M}ulti {L}ayer{P}erceptrons",
  author="Lodwich, Aleksander and Rangoni, Yves and Breuel, Thomas",
  booktitle="2009 {I}nternational {J}oint {C}onference on {N}eural {N}etworks",
  pages="1877--1884",
  year=2009,
  organization="IEEE"
)

@InProceedings(lukasik2020does,
  title ="Does label smoothing mitigate label noise?",
  author="Lukasik, Michal and Bhojanapalli, Srinadh and Menon, Aditya and Kumar, Sanjiv",
  booktitle ="Proceedings of the 37th International Conference on Machine Learning",
  pages="6448--6458",
  year=2020,
  editor="III, Hal Daumé and Singh, Aarti",
  volume=119,
  series="Proceedings of Machine Learning Research",
  month="13--18 Jul",
  publisher="PMLR",
  url="https://proceedings.mlr.press/v119/lukasik20a.html"
)

@MISC(Marquis2020Makerere,
  title    = "{B}ean {L}eaf {L}esions {C}lassification",
  author   = "Marquis03",
  year     =  2023,
  language = "en",
  howpublished = "Kaggle Repository",
  URL       = "https://www.kaggle.com/datasets/marquis03/bean-leaf-lesions-classification",
  note        = "Accessed: 03 January 2024"
)

@article(mondal2022adaptive,
  title="Adaptive {CNN} filter pruning using global importance metric",
  author="Mondal, Milton and Das, Bishshoy and Roy, Sumantra Dutta and Singh, Pushpendra and Lall, Brejesh and Joshi, Shiv Dutt",
  journal="Computer Vision and Image Understanding",
  volume=222,
  pages=103511,
  year=2022,
  publisher="Elsevier"
)

@article(moradi2020survey,
  title="A {S}urvey of {R}egularization {S}trategies for {D}eep {M}odels",
  author="Moradi, Reza and Berangi, Reza and Minaei, Behrouz",
  journal="{A}rtificial {I}ntelligence {R}eview",
  volume=53,
  pages="3947--3986",
  year=2020,
  publisher="{S}pringer"
)

@inproceedings(moreno2018forward,
  title="Forward {N}oise {A}djustment {S}cheme for {D}ata {A}ugmentation",
  author="Moreno-Barea, Francisco J and Strazzera, Fiammetta and Jerez, Jos{\'e} M and Urda, Daniel and Franco, Leonardo",
  booktitle="2018 IEEE {S}ymposium {S}eries on {C}omputational {I}ntelligence (SSCI)",
  pages="728--734",
  year=2018,
  organization="IEEE"
)

@article(morgan1989generalization,
  title="Generalization and {P}arameter {E}stimation in {F}eedforward {N}ets: Some {E}xperiments",
  author="Morgan, Nelson and Bourlard, Herv{\'e}",
  journal="{A}dvances in {N}eural {I}nformation {P}rocessing {S}ystems",
  volume=2,
  year=1989,
  pages="630 - 637"
)

@inproceedings(muller2019advances,
     author ="Rafael, Muller and Simon, Kornblith and Geoffrey E, Hinton",
     booktitle = "Advances in Neural Information Processing Systems",
     editor = "H. Wallach and H. Larochelle and A. Beygelzimer and F. d\textquotesingle Alch\'{e}-Buc and E. Fox and R. Garnett",
     publisher = "Curran Associates, Inc.",
     title = "When does label smoothing help?",
     url = "https://proceedings.neurips.cc/paper_files/paper/2019/file/f1748d6b0fd9d439f71450117eba2725-Paper.pdf",
     volume = 32,
     year = 2019,
    pages="4694 - 470",
    location = "Red Hook, NY, USA",
    note = "Accessed: 12 Decemder 2024"
)

@article(nakkiran2021deep,
  title="Deep {D}ouble {D}escent: Where {B}igger {M}odels and {M}ore {D}ata {H}urt",
  author="Nakkiran, Preetum and Kaplun, Gal and Bansal, Yamini and Yang, Tristan and Barak, Boaz and Sutskever, Ilya",
  journal="Journal of Statistical Mechanics: Theory and Experiment",
  volume=2021,
  number=12,
  pages=124003,
  year=2021,
  publisher="IOP Publishing"
)

@ARTICLE(nori2023effectiveness,
  author={Nori, Milad Khademi and Ge, Yiqun and Kim, Il-Min},
  journal={IEEE Access}, 
  title={On the Effectiveness of Activation Noise in Both Training and Inference for Generative Classifiers}, 
  year={2023},
  volume={11},
  number={},
  pages={131623-131638},
  doi={10.1109/ACCESS.2023.3335841}
)

@article(ohno2020auto,
  title="Auto-encoder-based {G}enerative {M}odels for {D}ata {A}ugmentation on {R}egression {P}roblems",
  author="Ohno, Hiroshi",
  journal="Soft Computing",
  volume=24,
  number=11,
  pages="7999--8009",
  year=2020,
  publisher="Springer"
)

@article(Olson2017PMLB,
    author="Olson, Randal S. and La Cava, William and Orzechowski, Patryk and Urbanowicz, Ryan J. and Moore, Jason H.",
    title="PMLB: a large benchmark suite for machine learning evaluation and comparison",
    journal="BioData Mining",
    year="2017",
    month="Dec",
    day="11",
    volume="10",
    number="36",
    pages="1--13",
    issn="1756-0381",
    doi="10.1186/s13040-017-0154-4"
)

@inproceedings(ozay2018training,
  title="Training {CNNs} with {N}ormalized {K}ernels",
  author="Ozay, Mete and Okatani, Takayuki",
  booktitle="Proceedings of the AAAI Conference on Artificial Intelligence",
  volume=32,
  number=1,
  year=2018,
pages="3884 - 3891"
)

@article(pan2009survey,
  title="A {S}urvey on {T}ransfer {L}earning",
  author="Pan, Sinno Jialin and Yang, Qiang",
  journal="IEEE Transactions on Knowledge and Data Engineering",
  volume=22,
  number=10,
  pages="1345--1359",
  year=2009,
  publisher="IEEE"
)

@inproceedings(petzka2021Relative,
	author = "Petzka, Henning and Kamp, Michael and Adilova, Linara and Sminchisescu, Cristian and Boley, Mario",
	booktitle = "Advances in Neural Information Processing Systems",
	editor = "M. Ranzato and A. Beygelzimer and Y. Dauphin and P.S. Liang and J. Wortman Vaughan",
	pages = "18420--18432",
	publisher = "Curran Associates, Inc.",
	title = "Relative Flatness and Generalization",
	volume = "34",
	year = "2021",
    location = "Red Hook, United States"
)

@misc(pereyra2017regularizing,
      title="Regularizing Neural Networks by Penalizing Confident Output Distributions", 
      author="Gabriel Pereyra and George Tucker and Jan Chorowski and Łukasz Kaiser and Geoffrey Hinton",
      year=2017,
      eprint="1701.06548",
      archivePrefix="arXiv",
      primaryClass="cs.NE",
      url="https://arxiv.org/abs/1701.06548", 
)

@MISC(Piosenka2023Types,
  title    = "30 {T}ypes of {B}alls {U}pdated {I}mage {C}lassification",
  author   = "Piosenka, Gerry",
  year     =  2023,
  language = "en",
  howpublished = "Kaggle Repository",
  URL       = "https://www.kaggle.com/datasets/samuelcortinhas/sports-balls-multiclass-image-classification",
  note        = "Accessed: 03 January 2024"
)

@article(prechelt1998automatic,
  title="Automatic {E}arly {S}topping {U}sing {C}ross {V}alidation: {Q}uantifying the {C}riteria",
  author="Prechelt, Lutz",
  journal="Neural {N}etworks",
  volume=11,
  number=4,
  pages="761--767",
  year=1998,
  publisher="Elsevier"
)

@inproceedings(reed2022self,
  title="Self-supervised {P}retraining {I}mproves {S}elf-supervised {P}retraining",
  author="Reed, Colorado J and Yue, Xiangyu and Nrusimha, Ani and Ebrahimi, Sayna and Vijaykumar, Vivek and Mao, Richard and Li, Bo and Zhang, Shanghang and Guillory, Devin and Metzger, Sean and others",
  booktitle="Proceedings of the IEEE/CVF Winter Conference on Applications of Computer Vision",
  pages="2584--2594",
  year=2022
)

@inproceedings(rifai2011higher,
  title="Higher {O}rder {C}ontractive {A}uto-encoder",
  author="Rifai, Salah and Mesnil, Gr{\'e}goire and Vincent, Pascal and Muller, Xavier and Bengio, Yoshua and Dauphin, Yann and Glorot, Xavier",
  booktitle="Machine Learning and Knowledge Discovery in Databases: European Conference, ECML PKDD 2011, Athens, Greece, September 5-9, 2011, Proceedings, Part II 22",
  pages="645--660",
  year=2011,
  location = "Berlin, Heidelberg",
  organization="Springer",
  location = "Cham, Switzerland"
)

@InProceedings(ronny2020Understanding,
  title="Understanding {G}eneralization {T}hrough {V}isualizations",
  author="Huang, W. Ronny and Emam, Zeyad and Goldblum, Micah and Fowl, Liam and Terry, Justin K. and Huang, Furong and Goldstein, Tom",
  booktitle="Proceedings on I Can't Believe It's Not Better! at NeurIPS Workshops",
  pages="87--97",
  year=2020,
  editor="Zosa Forde, Jessica and Ruiz, Francisco and Pradier, Melanie F. and Schein, Aaron",
  volume=137,
  series="Proceedings of Machine Learning Research",
  month="12 Dec",
  publisher="PMLR"
)

@inproceedings(rosenstein2005transfer,
  title="To {T}ransfer or {N}ot to {T}ransfer",
  author="Rosenstein, Michael T and Marx, Zvika and Kaelbling, Leslie Pack and Dietterich, Thomas G",
  booktitle="NIPS 2005 {W}orkshop on {T}ransfer {L}earning",
  volume=898,
  number=3,
  year=2005
)

@article(santos2022Avoiding,
    author = "Santos, Claudio Filipi Goncalves Dos and Papa, Joao Paulo",
    title = "Avoiding Overfitting: A Survey on Regularization Methods for Convolutional Neural Networks",
    year = 2022,
    issue_date = "January 2022",
    publisher = "Association for Computing Machinery",
    address = "New York, NY, USA",
    volume = 54,
    number = 10,
    issn = "0360-0300",
    url = "https://doi.org/10.1145/3510413",
    doi = "10.1145/3510413",
    pages = "1 - 25",
    journal = "ACM Comput. Surv.",
    month = "sep",
    articleno = 213,
    numpages = 25
)

@inproceedings(shin2018medical,
  title="Medical {I}mage {S}ynthesis for {D}ata {A}ugmentation and {A}nonymization {U}sing {G}enerative {A}dversarial {N}etworks",
  author="Shin, Hoo-Chang and Tenenholtz, Neil A and Rogers, Jameson K and Schwarz, Christopher G and Senjem, Matthew L and Gunter, Jeffrey L and Andriole, Katherine P and Michalski, Mark",
  booktitle="Simulation and Synthesis in Medical Imaging: Third International Workshop, SASHIMI 2018, Held in Conjunction with MICCAI 2018, Granada, Spain, September 16, 2018, Proceedings 3",
  pages="1--11",
  year=2018,
  organization="Springer",
  location = "Cham, Switzerland"
)

@article(shorten2019survey,
  title="A {S}urvey on {I}mage {D}ata {A}ugmentation for {D}eep {L}earning",
  author="Shorten, Connor and Khoshgoftaar, Taghi M",
  journal="Journal of Big Data",
  volume=6,
  number=1,
  pages="1--48",
  year=2019,
  publisher="SpringerOpen"
)

@misc(szegedy2015rethinking,
      title={Rethinking the Inception Architecture for Computer Vision}, 
      author="Christian, Szegedy and Vincent, Vanhoucke and Sergey, Ioffe and Jonathon, Shlens and Zbigniew, Wojna",
      year=2015,
      eprint="1512.00567",
      archivePrefix="arXiv",
      primaryClass="cs.CV",
      url="https://arxiv.org/abs/1512.00567" 
)

@InProceedings(tan20a,
  title="{D}rop{N}et: Reducing Neural Network Complexity via Iterative Pruning",
  author="Tan, Chong Min John and Motani, Mehul",
  booktitle="Proceedings of the 37th International Conference on Machine Learning",
  pages="9356--9366",
  year=2020,
  editor="III, Hal Daumé and Singh, Aarti",
  volume=119,
  series="Proceedings of Machine Learning Research",
  month="13--18 Jul",
  publisher="PMLR"
)

@article(tian2022comprehensive,
    title = "A comprehensive survey on regularization strategies in machine learning",
    journal = "Information Fusion",
    volume = "80",
    pages = "146-166",
    year = "2022",
    issn = "1566-2535",
    doi = "https://doi.org/10.1016/j.inffus.2021.11.005",
    author = "Yingjie Tian and Yuqi Zhang"
)

@inproceedings(trabucco2023effective,
    title="Effective {D}ata {A}ugmentation With {D}iffusion {M}odels", 
    author="Brandon Trabucco and Kyle Doherty and Max Gurinas and Ruslan Salakhutdinov",
    year=2023,
    eprint="2302.07944",
    archivePrefix="arXiv",
    primaryClass="cs.CV",
    url="https://arxiv.org/abs/2302.07944", 
)

@inproceedings(wiesler2014mean,
  title="Mean-normalized stochastic gradient for large-scale deep learning",
  author="Wiesler, Simon and Richard, Alexander and Schl{\"u}ter, Ralf and Ney, Hermann",
  booktitle="2014 IEEE International Conference on Acoustics, Speech and Signal Processing (ICASSP)",
  pages="180--184",
  year=2014,
  organization="IEEE"
)

@article(xu2023Comprehensive,
    title = "A Comprehensive Survey of Image Augmentation Techniques for Deep Learning",
    author = "Xu, Mingle and Yoon, Sook and Fuentes, Alvaro and Sun Park, Dong",
    journal = "Pattern Recognition",
    volume = "137",
    pages = "109347",
    year = "2023",
    issn = "0031-3203",
    doi = "https://doi.org/10.1016/j.patcog.2023.109347"
)

@inproceedings(yang2023image,
      title="{I}mage {D}ata {A}ugmentation for {D}eep {L}earning: {A} {S}urvey", 
      author="Suorong Yang and Weikang Xiao and Mengchen Zhang and Suhan Guo and Jian Zhao and Furao Shen",
      year=2023,
      eprint="2204.08610",
      archivePrefix="arXiv",
      primaryClass="cs.CV",
      url="https://arxiv.org/abs/2204.08610"
)

@ARTICLE(yangkun2023survey,
  author="Li, Yangkun and Ma, Weizhi and Chen, Chong and Zhang, Min and Liu, Yiqun and Ma, Shaoping and Yang, Yuekui",
  journal="IEEE Transactions on Knowledge and Data Engineering", 
  title="A Survey on Dropout Methods and Experimental Verification in Recommendation", 
  year=2023,
  volume=35,
  number=7,
  pages="6595-6615",
  doi="10.1109/TKDE.2022.3187013")

@inproceedings(yong2020gradient,
  title="Gradient {C}entralization: A {N}ew {O}ptimization {T}echnique for {D}eep {N}eural {N}etworks",
  author="Yong, Hongwei and Huang, Jianqiang and Hua, Xiansheng and Zhang, Lei",
  booktitle="Computer Vision--ECCV 2020: 16th European Conference, Glasgow, UK, August 23--28, 2020, Proceedings, Part I 16",
  pages="635--652",
  year=2020,
  location = "Cham, Switzerland",
  organization="Springer"
)

@article(zhang2021understanding,
  title="Understanding {D}eep {L}earning ({S}till) {R}equires {R}ethinking {G}eneralization",
  author="Zhang, Chiyuan and Bengio, Samy and Hardt, Moritz and Recht, Benjamin and Vinyals, Oriol",
  journal="Communications of the ACM",
  volume=64,
  number=3,
  pages="107--115",
  year=2021,
  publisher="ACM New York, NY, USA"
)

@article(zhang2018mixup,
      title="Mixup: {B}eyond {E}mpirical {R}isk {M}inimization", 
      author="Hongyi Zhang and Moustapha Cisse and Yann N. Dauphin and David Lopez-Paz",
      year=2018,
      eprint="1710.09412",
      archivePrefix="arXiv",
      primaryClass="cs.LG",
      url="https://arxiv.org/abs/1710.09412"
)

@inproceedings(zhou2019toward,
  title="Toward {U}nderstanding the {I}mportance of {N}oise in {T}raining {N}eural {N}etworks",
  author="Zhou, Mo and Liu, Tianyi and Li, Yan and Lin, Dachao and Zhou, Enlu and Zhao, Tuo",
  booktitle="International Conference on Machine Learning",
  pages="7594--7602",
  year=2019,
  organization="PMLR"
)

@inproceedings(zhuang2015supervised,
  title="Supervised {R}epresentation {L}earning: Transfer {L}earning with {D}eep {A}utoencoders",
  author="Zhuang, Fuzhen and Cheng, Xiaohu and Luo, Ping and Pan, Sinno Jialin and He, Qing",
  booktitle="Twenty-fourth International Joint Conference on Artificial Intelligence",
  year=2015,
    pages="4119 - 4125"
)

@article(zhuang2020comprehensive,
  title="A {C}omprehensive {S}urvey on {T}ransfer {L}earning",
  author="Zhuang, Fuzhen and Qi, Zhiyuan and Duan, Keyu and Xi, Dongbo and Zhu, Yongchun and Zhu, Hengshu and Xiong, Hui and He, Qing",
  journal="Proceedings of the IEEE",
  volume=109,
  number=1,
  pages="43--76",
  year=2020,
  publisher="IEEE"
)

@article(khatami2020weight,
  title={A weight perturbation-based regularisation technique for convolutional neural networks and the application in medical imaging},
  author={Khatami, Amin and Nazari, Asef and Khosravi, Abbas and Lim, Chee Peng and Nahavandi, Saeid},
  journal={Expert systems with applications},
  volume={149},
  pages={113196},
  year={2020},
  publisher={Elsevier}
)

@inproceedings(yang2021comparative,
  title={Comparative analysis of structured pruning and unstructured pruning},
  author={Yang, Zhengwu and Zhang, Han},
  booktitle={International Conference on Frontier Computing},
  pages={882--889},
  year={2021},
  organization={Springer},
location = {Cham, Switzerland}
)

@article(he2023structured,
  title={Structured pruning for deep convolutional neural networks: A survey},
  author={He, Yang and Xiao, Lingao},
  journal={IEEE transactions on pattern analysis and machine intelligence},
  volume={46},
  number={5},
  pages={2900--2919},
  year={2023},
  publisher={IEEE}
)

@ARTICLE{awais2025,
  author={Awais, Muhammad and Naseer, Muzammal and Khan, Salman and Anwer, Rao Muhammad and Cholakkal, Hisham and Shah, Mubarak and Yang, Ming-Hsuan and Khan, Fahad Shahbaz},
  journal={IEEE Transactions on Pattern Analysis and Machine Intelligence}, 
  title={Foundation Models Defining a New Era in Vision: A Survey and Outlook}, 
  year={2025},
  volume={47},
  number={4},
  pages={2245-2264},
  doi={10.1109/TPAMI.2024.3506283}}

@article{zhou2024comprehensive,
  title={A comprehensive survey on pretrained foundation models: A history from bert to chatgpt},
  author={Zhou, Ce and Li, Qian and Li, Chen and Yu, Jun and Liu, Yixin and Wang, Guangjing and Zhang, Kai and Ji, Cheng and Yan, Qiben and He, Lifang and others},
  journal={International Journal of Machine Learning and Cybernetics},
  pages={1--65},
  year={2024},
  publisher={Springer},
  volume = {16}
}

@inproceedings{bosman2020loss,
  title={Loss surface modality of feed-forward neural network architectures},
  author={Bosman, Anna Sergeevna and Engelbrecht, Andries Petrus and Helbig, Mard{\'e}},
  booktitle={2020 International Joint Conference on Neural Networks (IJCNN)},
  pages={1--8},
  year={2020},
  organization={IEEE}
}
\begin{IEEEbiography}[{\includegraphics[width=2.8cm]{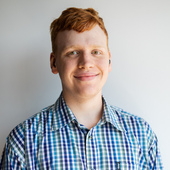}}]{Christiaan P. Opperman} is a Masters student in Computer Science at the Department of Computer Science, University of Pretoria, South Africa. He is a software engineer with experience in an array of different areas, ranging from designing systems to data science at an international software development firm (Flatrock Solutions). He is interested in the generalisation ability of neural networks, in understanding how neural networks work, in fly fishing, and in photography.

\end{IEEEbiography}
\vspace{0.5cm}
\begin{IEEEbiography}[{\includegraphics[width=2.8cm]{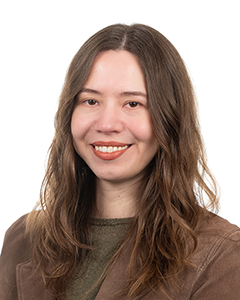}}]{Anna S. Bosman } is currently appointed as an associate professor in the Department of
Computer Science, University of Pretoria, South Africa, where she received the Ph.D.
degree in Computer Science in 2019. Her research interests include deep neural networks, loss landscape analysis, energy-efficient models, meta-learning, and computer vision applications. She has been a member of both the IEEE and the IEEE Computational Intelligence Society (CIS) since 2014, and a member of the IEEE CIS Neural Networks Technical Committee (NNTC) since 2023.
\end{IEEEbiography}

\vspace{0.5cm}
\begin{IEEEbiography}[{\includegraphics[width=2.8cm]{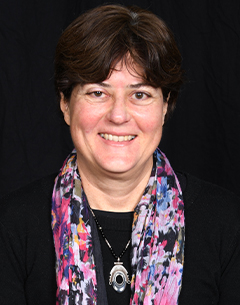}}]{Katherine M. Malan}{\space} is a full professor in the Department of Decision Sciences at the University of South Africa. She received her PhD in computer science from the University of Pretoria in 2014 and her MSc \& BSc degrees from the University of Cape Town. She has 30 years' lecturing experience, mostly in Computer Science, at three different South African universities. Her research interests include automated algorithm / model selection in optimisation and machine learning, fitness landscape analysis and the application of computational intelligence techniques to real-world problems. She serves as editor-in-chief of the South African Computer Journal, associate editor for Engineering Applications of Artificial Intelligence and chair of the IEEE Evolutionary Computation Technical Committee (2024 - 2025).

\end{IEEEbiography}

\end{document}